\documentclass[journal,twocoulmn,twoside]{IEEEtran}    % TWOCOLS
\usepackage{booktabs}
\usepackage{multirow}
\usepackage{amsfonts}
\usepackage{amssymb}
\usepackage{amsbsy} % Per a \boldsymbol
\usepackage{amsmath}
\usepackage[ansinew]{inputenc}%
\usepackage{cite}
\usepackage{multirow}
\usepackage{color}
\usepackage{lscape}
\usepackage[dvips]{graphicx}
\usepackage{colortbl}
\usepackage{url}
\usepackage{algorithm}
\usepackage{algorithmic}
\usepackage{stmaryrd}
\usepackage{hyperref}
\newcommand{\figdir}{figs_paper/}

\def\w{\ensuremath{\mathbf{w}}}

\def\x{\ensuremath{\mathbf{x}}}

\DeclareMathOperator*{\argmax}{argmax}

\newcommand{\mytitle}{Dense semantic labeling of sub-decimeter resolution images with convolutional neural networks}

\begin{document}

\title{\mytitle}
\author{Michele~Volpi,~\IEEEmembership{Member,~IEEE},~Devis~Tuia,~\IEEEmembership{Senior Member,~IEEE \vspace*{-5mm}}
\thanks{Manuscript received YYXX, 2016;} 
\thanks{\noindent MV and DT are with the MultiModal Remote Sensing group, University of Zurich, Switzerland. Email: \{michele.volpi,devis.tuia\}@geo.uzh.ch, web: http://geo.uzh.ch/ , Phone: +4144 635 51 11, Fax: +4144 635 68 48.}
\thanks{\noindent Digital Object Identifier xxxx}}

%\markboth{IEEE Transactions on Geoscience and Remote Sensing, 2016}{Volpi et al.: \mytitle}
\markboth{}{Volpi et al.: \mytitle}

\maketitle

%\newpage

\begin{abstract}
Semantic labeling (or pixel-level land-cover classification) in ultra high resolution imagery \mbox{($<$ 10cm)} requires statistical models able to learn high level concepts from spatial data, with large appearance variations. Convolutional Neural Networks (CNNs) achieve this goal by learning discriminatively a hierarchy of representations of increasing abstraction. 

In this paper we present a CNN-based system relying on an downsample-then-upsample architecture. Specifically, it first learns a rough spatial map of high-level representations by means of convolutions and then learns to upsample them back to the original resolution by deconvolutions. By doing so, the CNN learns to densely label every pixel at the original resolution of the image. This results in many advantages, including i) state-of-the-art numerical accuracy, ii) improved geometric accuracy of predictions and iii) high efficiency at inference time.%, since the final pixelwise prediction can be obtained by a single forward pass with no loss in spatial accuracy, i.e. the output has exactly the same size as the input.

We test the proposed system on the Vaihingen and Potsdam sub-decimeter resolution datasets, involving semantic labeling of aerial images of 9cm and 5cm resolution, respectively. These datasets are composed by many large and fully annotated tiles allowing an unbiased evaluation of models making use of spatial information. We do so by comparing two standard CNN architectures to the proposed one: standard patch classification, prediction of local label patches by employing only convolutions and full patch labeling by employing deconvolutions. All the systems compare favorably or outperform a state-of-the-art baseline relying on superpixels and powerful appearance descriptors. The proposed full patch labeling CNN outperforms these models by a large margin, also showing a very appealing inference time.
\end{abstract}

\IEEEpeerreviewmaketitle

% !TEX root = main-volpi-cnn-deconv-journal.tex
\section{Introduction}\label{sec:intro}

\IEEEPARstart{S}{emantic} labeling is the task of assigning a semantic label (land-cover or land-use class) to every pixel of an image. When processing ultra-high resolution data, most of state-of-the-art methods rely on supervised classifiers trained on specifically hand-crafted feature sets (appearance descriptors), describing locally the image content. The extracted high-dimensional representation is assumed to contain enough information to cope with the ambiguities caused by the limited spectral information of the ultra-high resolution sensors.

%% standard state-of-the-art refs about spatial feature extraction + segmentation (after 2013-2014 I would say)
In the pipeline described above, input images undergo a spatial feature extraction step implemented by specific operators on local portions of the image (patches, superpixels or regions, objects, etc.), so that particular \emph{spatial} arrangements of colors are encoded into a high-dimensional representation. A supervised classifier is usually employed to learn a mapping from the appearance descriptors to the semantic label space, which in turn allows to assign a label to every region of a previously unseen test image.
%
%This is often referred to as \emph{spatial information} and is often used in conjunction with spectral information to define the \emph{input space}. Then, a supervised classifier employs this high-dimensional representation to learn a mapping to the \emph{output space}, allowing to predict semantic class labels on unseen test data. %While approaching semantic labeling problems, it is common to extract dense features from the whole image space. These features provide one different value for each pixel to be classified. 
Common examples of spatial features are texture statistics, mathematical morphology and oriented gradients \cite{fauvel2013pieee}. Other common approaches strategies rely on bag-of-visual-words \cite{sivic2003cvpr}. This mid-level representation is based on a quantization of appearance descriptors such as gradients, orientations, texture (usually obtained with a clustering algorithm). This quantization is then pooled spatially into histograms of spatial occurrences of cluster labels, or bag-of-visual-words. For instance, in \cite{gueguen2015tgrs} bag-of-visual-words are used to classify image tiles and detect compound structures.

%% Feature learning, general in RS -> motivate deep learning
The drawback of these approaches is that the features depend on a specific (set of) feature extraction method, whose performance on the specific data is a-priori unknown. Moreover, most appearance descriptors depend on a set of free parameters, which are commonly set by user experience via experimental trial-and-error or cross-validation \cite{fauvel2013pieee,shotton2006eccv}. Exhaustive and global optimization of such values is unfeasible in reasonable time, but the selection from random feature ensembles has shown to be an effective proxy \cite{tuia2014tgrs,tokarczyk2014tgrs,tuia2015jisprs}. %These latter approaches rely on the response of a supervised classifier, which drives the selection of the specific filter family. So, even if 
In these cases, the filter families from which to chose from are predefined and the parameters of the system are selected heuristically by random search to minimize the error over the semantic labeling task. Although the selection of features is data-driven, the filters themselves are still not learned end-to-end from the data, thus potentially sub-optimal.
  
%% General intro to CNN (avoid intro on deep learning though)
Deep learning deals with the development of systems trainable in an end-to-end fashion. End-to-end usually means learning jointly a series of feature extraction from raw input data to a final, task-specific, output. %\blue{so that the task on which the system is trained on provides low generalization error. Although the systems are rarely explicitly trained for ``representation learning'', but rather on cost functions for which computing gradients is relatively easy and allows fast learning, such as denoising, classification or regression.} 
%Those features correspond to generalizations of the inputs related to the output to be predicted and are of higher abstraction as long as we go deeper in the network.
All deep learning systems implicitly learn representations optimizing the loss on top of the network, driving the training of the model's weights. They usually minimize a task-specific differentiable loss function for classification, regression, semantic labeling, super-resolution or depth estimation, and the network learns representations which minimize such loss. %In the specific case of CNNs, filters learned in first layers have been shown to be generic image descriptors, shared across many tasks \cite{yosinski2014nips}.} %But irrespectively of the network architecture and loss used, the resulting learned representations will correspond to concepts of increasing abstraction thanks to the hierarchical and sequential processing of most networks. 
% In principle, the more complicated the relationship between inputs and outputs, the more complex the representation should be, to better depict input-output relationships.
Most common deep learning algorithms are (stacked) autoencoders \cite{bengio2006nips,vincent2008icml,kavukcuoglu2010nips}, restricted Boltzmann machines \cite{nair2010icml,mnih2012icml} and deep belief networks \cite{hinton2006science,bengio2009deepai}. For a review of main approaches we refer to \cite{bengio2009deepai,goodfellow2016deepbook}. In this paper we focus on Convolutional Neural Networks (CNNs)~\cite{lecun1998pieee}. Differently from other approaches, CNNs were specifically designed for image classification tasks, i.e. assigning a single class label to an entire image / scene. Representations are obtained by learning a hierarchy of convolution filters from the raw image. All the weights of the convolutions are learned end-to-end to minimize the classification error of the model.

%The rationale behind CNN is that a complete system composed by spatial feature extraction and classification can be learned jointly, by minimizing a loss function by backpropagation and iteratively updating (i.e. learning) convolutional filters weights. Moreover, some layers are designed to enforce invariance to local translations and small rotations, making the system an excellent predictors with very low generalziation error. %The fact that CNN share a predefined number of filters at each layer makes them trainable even in setups where many layers are stacked (very deep nets), making them a nonlinear classifier with very high capacity. Modern CNN are trained in a fully supervised way, in contrast proposed in the mid 2000's, which were mostly unsupervised systems (or initialized as such).

%% CNN in vision propblems (classif)
CNNs have become extremely successful in many modern, high-level, computer vision tasks, ranging from image classification to object detection, depth estimation and semantic labeling. First examples of deep CNN architectures have been proposed for image classification problems. The most notable example is the ILSVRC challenge in 2012\footnote{Image Large Scale Visual Recognition Challenge, \url{http://www.image-net.org/challenges/LSVRC/}}, where CNNs significantly outperformed the state-of-the-art systems based on handcrafted appearance descriptors \cite{krizhevsky2012nips}. Notable extensions allowing to train deeper CNN (i.e. adding trainable layers and thus increasing the capacity of the model) \cite{simonyan2015iclr,szegedy2015cvpr,he2015cvpr} were the introduction of drop-out\cite{srivastava2014jmlr}, batch normalization \cite{ioffe2015icml} and other strategies allowing better propagation of gradients, such as rectified linear units (ReLU) nonlinearities \cite{jarrett2009iccv,he2015cvpr}. Together with the (very) large annotated training datasets and powerful GPU making it possible to train such models, these intuitions made CNNs the gold standard for image classification problems. 

%% Segmentation
CNNs have been also adapted for semantic labeling (semantic segmentation) problems \cite{farabet2013pami,girshick2014cvpr,long2015cvpr,noh2015iccv}. These papers show two distinct approaches: %: one is based on learning region (or object) representations, while the other aims at learning directly pixel-level labelings.
In the first case, the model is trained to predict a single class label for each region (patch, superpixel or object proposal). The output is usually a vector of scores or probabilities for each class, based on the appearance of an entire region. %The CNN architecture learns convolutions and downsamples the input patch signal (by strided convolutions and poolings), so that hierarchical and nonlinear feature representations are learned by the model.
In the second case, the network is trained to predict \emph{spatial  arrangements} of labels at pixel-level. These architectures are able to model local structures (e.g. spatial extent, class co-occurrences) across the input space. These \emph{upsampling} steps are formulated by means of \emph{deconvolutions} \cite{long2015cvpr,noh2015iccv}. In this paper, we adopt this approach and propose a strategy to learn locally dense semantic labeling of patches.

\subsection{Deep learning in remote sensing} 

% Image classification
Remote sensing image processing pipelines are beginning to exploit deep learning. Initially, such systems tackled the problem of image / tile classification: i.e. assigning to large patches (or small images) a single semantic label such as ``urban'', ``sparse urban'' or ``forest''. This task is well represented by the UC Merced landuse classification  dataset\footnote{\url{http://vision.ucmerced.edu/datasets/landuse.html}}. Marmanis {\it et al.} \cite{marmanis2015grsl} present a two-stages system, in which pretrained CNNs are combined with a second stage of supervised training. This strategy mitigates overfitting and shows excellent performances. Penatti {\it et al.} \cite{penatti2015cvprw} show that models pre-trained on general image classification tasks (specifically on ILSVRC) can be used as generic feature extractors also for remote sensing image (tile) classification, outperforming most of the state-of-the-art feature descriptors. Authors point out that such models were particularly well suited for high resolution aerial data. Finally, Castelluccio {\it et al.} \cite{castelluccio2015arxiv} also explore pretrained architectures. They also study the impact of domain adaptation by fine tuning the model to the new task (from ILSVRC data to remote sensing images) or by training from scratch, always keeping exactly the same CNN architectures. They show that pre-trained models fine-tuned on remote sensing data perform better than models with a same architecture but trained from scratch (with randomly initialized weights). This indicates that CNN architectures devoted to natural image classification tasks without specific adjustments tend to overfit remote sensing data. Typically, to model complex appearance variations, CNNs devoted for image classification contain many parameters, particularly in the fully connected layers, which could easily contain more than 90\% (VGG network \cite{simonyan2015iclr}) or 95\% (AlexNet \cite{krizhevsky2012nips}) of the total number of learnable parameters. With the typical remote sensing datasets such large number of parameter would be hard to estimate, because of the limited amount of labeled samples they generally provide. For this reason, training networks for remote sensing problems requires adjustments.

\begin{figure}
\centering
\includegraphics[width=.45\textwidth]{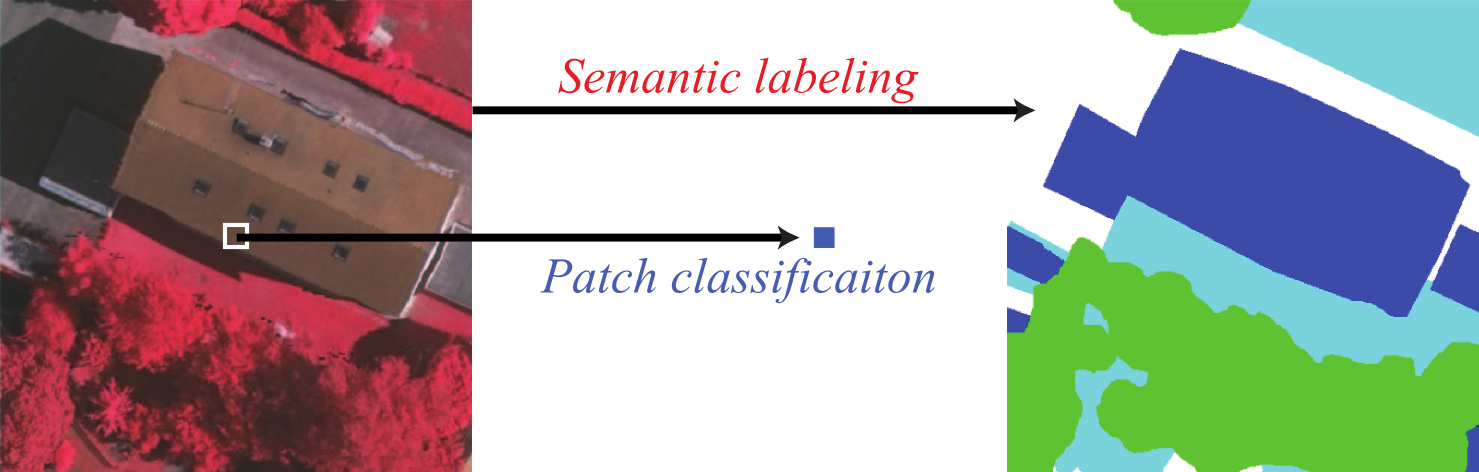}
\caption{Comparing patch classification and semantic labeling: the first learns a single label per patch (assumed to be the one of the central pixel), while the second learns to densely predict semantic labels for each location. \label{fig:patchvssem}}
\vspace*{-4mm}
\end{figure}

% dense labeling
Recently, semantic labeling tasks in remotely sensed data were also approached by means of deep learning. Chen {\it et al.} \cite{chen2014jstars} rely on stacked autoencoders, trained to reconstruct PCA-compressed hyperspectral signals. The network is then fine-tuned by backpropagating errors from a softmax loss on top of the stacked autoencoders, which also provides the final classification of the pixels. Firat {\it et al.} \cite{Firat2014icpr} train a sparse convolutional autoencoder to perform object detection in remote sensing images. Although the model is only composed of a (wide) single layer and technically is not ``deep'', the idea of representation learning is tackled elegantly. Zhang {\it et al.}  \cite{zhang2015tgrs} propose a system based on stacked autoencoders: rather than training the model on random patches from the images, they follow a strategy sampling patches based on visual saliency, shown to improve over the model without region selection. Romero {\it et al.} \cite{romero2016tgrs} propose a deep convolutional sparse autoencoder relying on a specific sparsity criterion. Generic features are extracted for image patches and a separate classification is then performed to label each feature vector. These works have the attractivity of working in scarcely supervised settings, by employing unsupervised (pre-)training schemes. However, these models can not learn discriminative representations, which is a task usually left to a classifier trained by using such generic representations. Authors in \cite{paisitkriangkrai2015cvprw} propose a system based on several CNNs trained on the Vaihingen challenge dataset (see Section~\ref{sec:data} for the description of the dataset) to perform semantic labeling. The potential of CNNs is clearly shown by combining the features extracted from the CNN with random forest classifiers, standard appearance descriptors and conditional random fields performing structured prediction on the probabilities given by the classifier. However, the CNN in \cite{paisitkriangkrai2015cvprw} are not trained specifically for pixel-wise semantic labeling tasks, but rather for patch classification: the network is designed to predict a single label from a patch (independently from the others, as in the ``patch classification'' part of Fig.~\ref{fig:patchvssem}). Sherrah \cite{sherrah2016arxiv} proposed a no-downsampling system relying on pretrained networks, together with a network designed to process digital surface model data only, to predict class-conditional scores. Fused results are then post-processed by a conditional random field. So far, these results are among the best on the benchmark data.

Training a network explicitly for semantic labeling or for classification problems are very different task, as illustrated in Fig.~\ref{fig:patchvssem}. In the former setting, we would like to train a model that is able to label each pixel present in the image. This should be achieved not only by learning the relationship between colors and labels, but mostly by learning and taking into account spatial relationships at different scales. %: we call this setting, which lies the labeling of groups of pixels simultaneously, semantic labeling. %In such setting, predicting the label of groups of pixels simultaneously and by labeling all the required pixels we are in the setting of semantic labeling. 
In the second setting (patch classification), a network trained for classification predicts a single label per patch. This patch-level label is then assumed to be the label of the central pixel. In this setting, many queries are spatially concatenated to obtain the prediction map. In addition to being extremely inefficient, patch classification-based strategies are inappropriate, since they do not explicitly learn spatial configuration of labels and consequently oversmooth objects boundaries. To make an example, a CNN model aiming at classifying a patch centered on a car lying on a road could score high for both the classes ``road'' and ``car'':  both solutions are semantically correct, since both classes are somehow lying in the center of the patch. Given the content and the context, it's hard to penalize one solution more than the other. In the semantic labeling case, we would like to predict the label of \emph{each pixel} in the patch \emph{jointly}, thus avoiding ambiguities given by the content of the patch and de facto performing structured prediction by learning class-specific structures contained in each patch.

%% Our strategy's motivations
In this paper, we explicitly tackle the semantic labeling problem and we show how a modern CNN-based system can be trained for dense semantic labeling tasks in a fully supervised fashion. We are particularly interested in the semantic labeling of ultra-high spatial resolution images, where data contains a tremendous amount of geometrical information, but with a limited number of spectral channels. This is typically the case with off-the-shelf UAV and most aerial imaging systems. % and many very-high resolution satellites such as WorldView or Pl{\'e}iades. 
%We propose a methodology developed specifically for the semantic labeling of remote sensing images that takes advantage of the spatial structuring of very to ultra-high resolution images without relying on pre-trained architectures coming from other fields of study. 
We introduce a patch-based deconvolution network to first encode land-cover representations in a rough spatial map (that we name bottleneck) and then to upsample them back to the original input patch size. The modeling power of this downsampling-then-upsampling architecture relies on the fact that global spatial relationships can be modeled directly by learning locally, on a coarser spatial signal (downsampling part). The upsampling will then take into account local spatial structuring for each class, while extracting nonlinear representations at the same time. Training the network patch-wise allows us to deal with images of any size, by decomposing the problem into sub-regions with representative spatial support. The structure of the network is given in Fig.~\ref{fig:fullnet}. To train it, we employ standard stochastic gradient descent with momentum, on batches of training patches sampled from the training images. At inference time, we again take a step away from standard approaches that must crop large images into densely overlapping blocks (i.e. with a small stride) or rely on object proposals / regions) to maximally preserve the spatial resolution of predictions. In our system, the whole image can be directly fed to the trained network to obtain a posterior probability map of semantic labels \emph{without loss in resolution}. Doing the same with standard patch-based CNNs would show a drastic loss in spatial resolution.

Differently from already published works in remote sensing, we train the network specifically for dense labeling, and not for classification. Compared to the no-downsampling extension of \cite{sherrah2016arxiv}, we argue that a downsample-then-upsample architecture could make better use of contextual relationships, since without downsampling activations are location specific and information is not explicitly shared across scales (layers) if not via learned filters. Results, however, suggest that both strategies seem appropriate. The main difference between our work and previous computer vision studies \cite{long2015cvpr,noh2015iccv} relies in the fact that our upsampling layers (deconvolution) learn spatial filter by an initial, coarser spatial map of activations (bottleneck layer). Instead, \cite{noh2015iccv} perform upsampling from a single feature vector, by exactly mirroring the downsampling layers (thus not learning deconvolution filters with increasing dimensionality and consequent expressive power) and propagating pooling activations for unpooling signals. Moreover, to delineate ambiguous areas, they also make use of object proposals at test time. In \cite{long2015cvpr} the activations from lower layers are combined with high-stride upsamplings, to cope oversmoothing. In our approach, the CNN directly delineates classes accurately without the need of relating different layers outputs.%, such as combining object proposals at inference time \cite{noh2015iccv}.}% We found that the proposed solution is geometrically accurate, even without requiring object class proposals to obtain the final predictions, as in~\cite{noh2015iccv}.

%We detail each step and provide comprehensive overview about how to set up such architecture. %, depending on pooling sizes and strides, making such results too coarse. %We detail our training strategy and we release our model as a MatConvNet net file\footnote{} on our website.

%\begin{figure*}[!t]
%  \includegraphics[width=\textwidth]{\figdir net-fdeconv-lowres.png}
%\vspace*{-4mm}
%\caption{Schematic of the network proposed to perform dense patch-based semantic labeling. %\label{fig:fullnet}}
%\end{figure*}

% Summary of the paper / TOC done the clever way
Summing up, we propose to train a CNN for semantic labeling tasks by employing a dowsampling-then-upsampling architecture. We first review main blocks of CNN architectures (summarized in Sec.~\ref{sec:models}), which allows us to carefully review differences between the CNN we put forward and more standard strategies in Sec.~\ref{sec:ourmodel}, where we also provide important details to set up such systems. We test our intuitions on two challenging aerial images dataset, the recently released Vaihingen and Potsdam semantic labeling challenges (presented in Sec.~\ref{sec:resu}). Here, we also show how the proposed method compares to standard CNN approaches and we discuss its strengths with respect to the state-of-the-art. In Sec.~\ref{sec:concs} we conclude the paper by summarizing main contributions.

%%% Local Variables: 
%%% mode: latex
%%% TeX-master: "main-volpi-cnn-deconv-journal"
%%% End:

% !TEX root = main-volpi-cnn-deconv-journal.tex
\renewcommand{\tabcolsep}{1pt}

\section{Convolutional Neural Networks}\label{sec:models}

CNNs are composed by a sequential hierarchy of processing layers (Fig.~\ref{fig:fullnet}). From the input to the final classification layer, data go trough a series of trainable units. A general feed forward network can bee seen as a concatenation of functions, starting from some input $\x$: 
\begin{equation}
  g(\x) = g_L\bigg(g_{L-1}\Big(g_{l}\big(g_1\left(\x;\w_1\right);\w_{l}\big);\w_{L-1}\Big);\w_L\bigg). 
\end{equation}
The functions $g_l$ composing the $L$ layers of the network $g$ are usually linear functions, subsequently passed through nonlinearities, while weights $\w_l$ are learned from data. For instance, multilayer perceptrons model input-output relationships by a series of densely interconnected hidden layers composed by linear units and nonlinear activations functions. CNN are structured in a similar way, but neurons are learnable convolutions shared at each image location.

In the following, we provide a comprehensive introduction to the structure of our CNN and present important strategies to reduce overfitting while training.

\subsection{CNN building blocks}

\begin{figure}[!t]
\centering
\begin{tabular}{rl} 
\raisebox{1mm}{\rotatebox{90}{\small CNN-PC}} & \includegraphics[height=1.56cm]{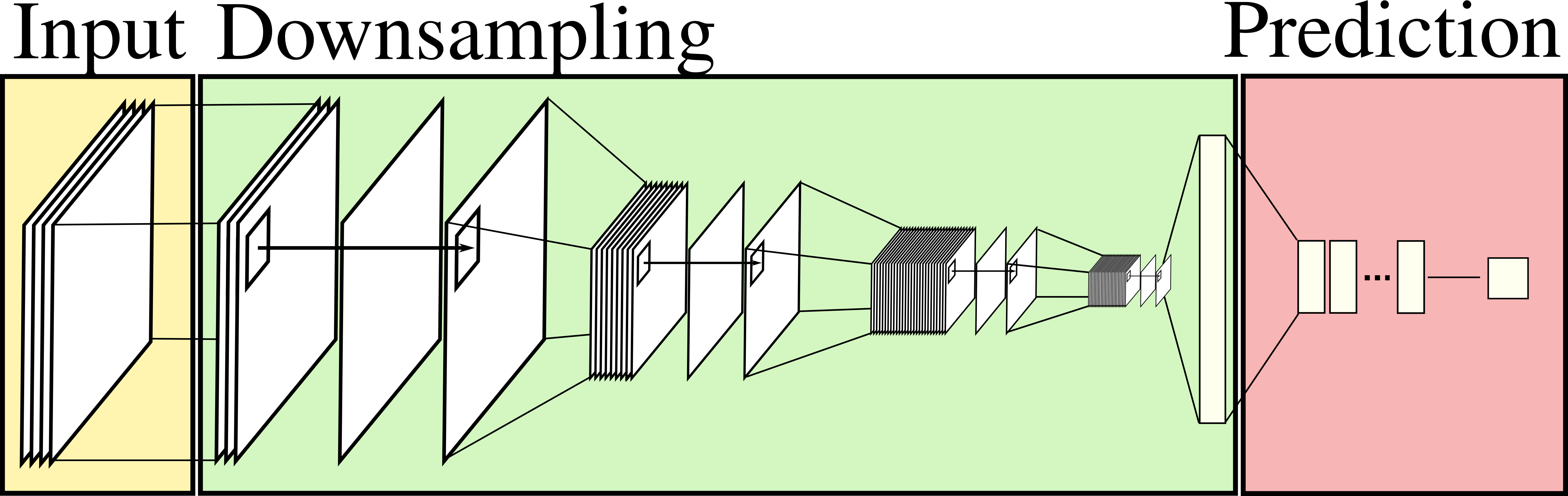} \\
\raisebox{1mm}{\rotatebox{90}{\small CNN-SPL}} & \includegraphics[height=1.56cm]{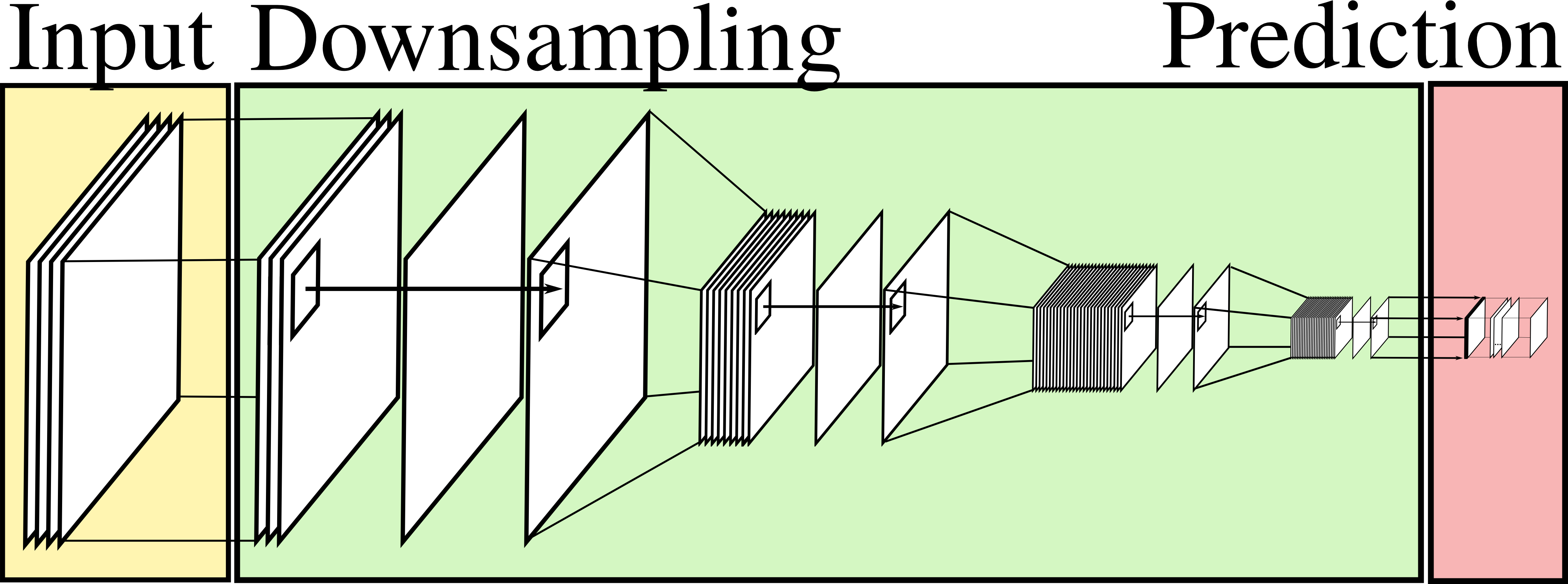}\\
\raisebox{1mm}{\rotatebox{90}{\small CNN-FPL}} & \includegraphics[height=1.56cm]{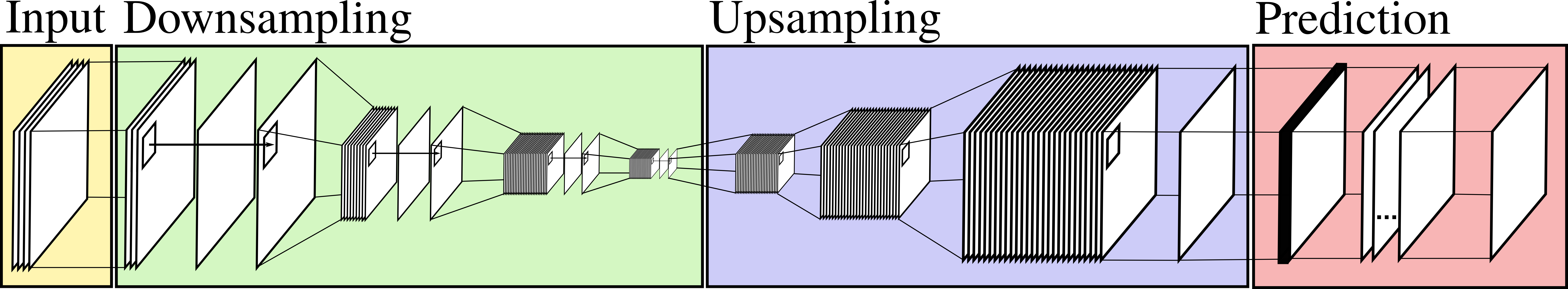} \\
\end{tabular}
\vspace*{-4mm}
\caption{Schematic of the networks used to perform semantic labeling. \label{fig:fullnet}}
\vspace*{-4mm}
\end{figure}

In this section, we detail five building blocks of the CNN architectures this paper deals with: convolutions, nonlinear activations, spatial pooling, deconvolutions and loss function. In Fig.~\ref{fig:fullnet}, schematics of the CNNs are illustrated. Note that only the proposed CNN full patch labeling (CNN-FPL) makes use of deconvolutions, while the CNN patch classification (CNN-PC) and CNN semi patch labeling (CNN-SPL) use standard blocks. In the following, we refer to the inputs and outputs to the $l$th layer as $\x^l$ and $\x'^l = \x^{l+1}$, respectively. We will simply refer to $\x$ and $\x'$, since the layer indexing is clear from the context.

\paragraph{Convolutions}

The main building blocks of a CNN are convolutional layers. A convolutional layer is a set of $K'$ filters (or neurons) with learnable parameters $\w$. In each neuron, parameters $\w$ are arranged in an array of size $M \times M \times K$ to process a $K$ dimensional input. For instance, in RGB data $K = 3$ in the first convolutional layer. Therefore, $K'$ 3-dimensional $M \times M$ filters map to $K'$-dimensional activations.

\begin{figure}[!t]
\centering
\begin{tabular}{cc}
\includegraphics[width=0.23\textwidth]{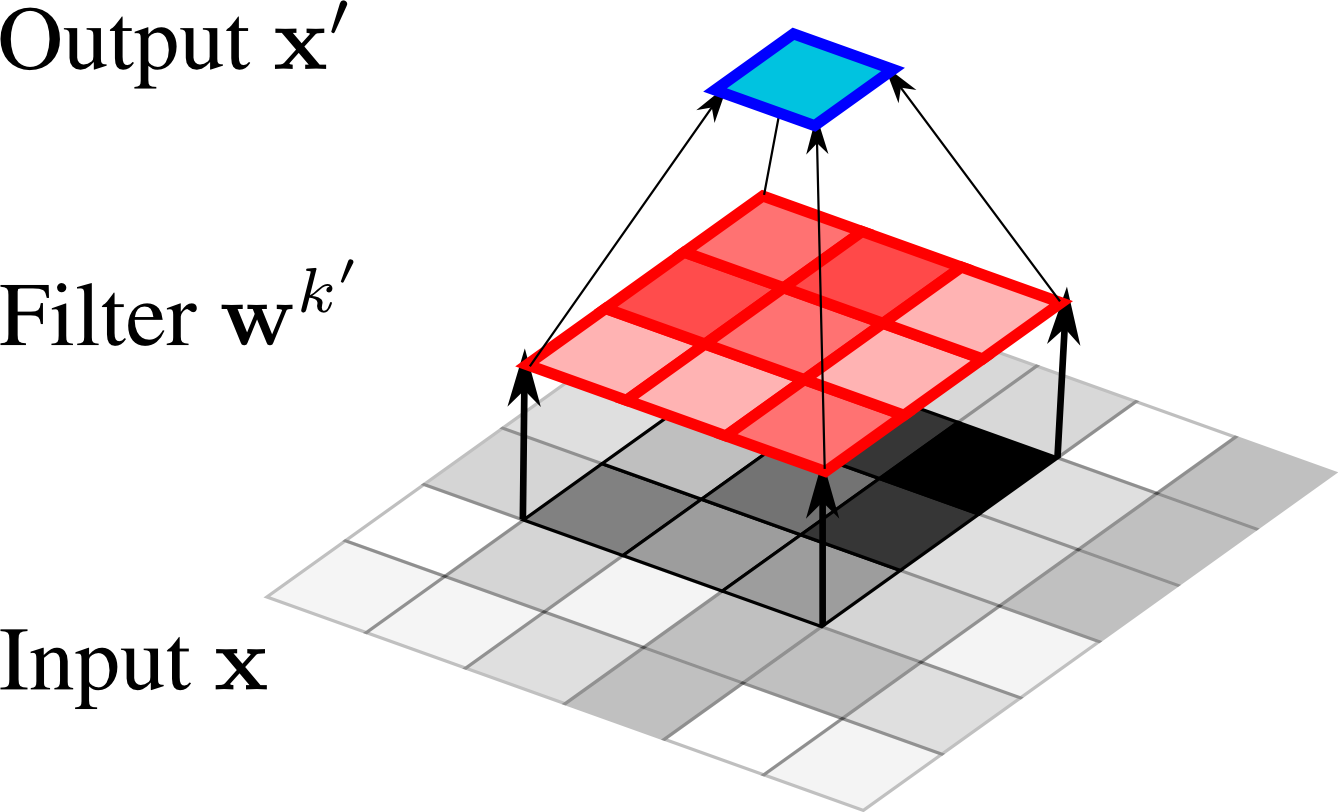} & % was 0.23
\includegraphics[width=0.23\textwidth]{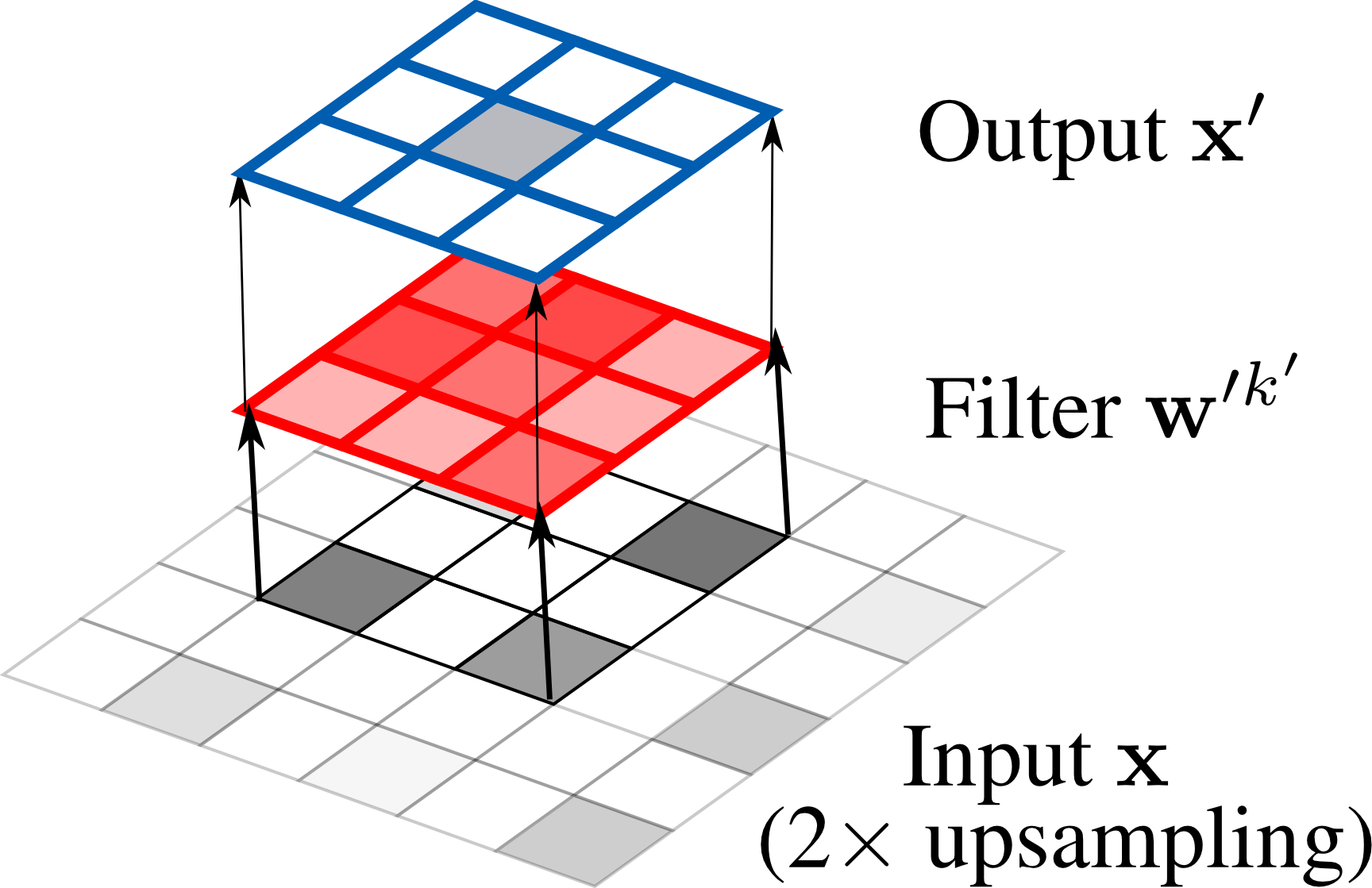} \\
(a) & (b)
\end{tabular}
\caption{Example of (a) convolution and (b) deconvolution. \label{fig:conv}}
\vspace*{-4mm}
\end{figure}

We center neurons on $i$ and $j$, denoting spatial coordinates relative to its input. Thus, the response for the $k'$-th filter is:
\begin{equation}
\x'_{ijk'} = \sum^K_{k=1} \sum^M_{q=1} \sum^M_{p=1} \w_{pqk}\cdot\x_{pqk} + b,
\end{equation}
where $b$ is a learned bias term. Supposing we have an input image of size $N\times L\times K$, the output of the convolutional layer is $\left(\frac{(N-M + 2z)}{s} + 1\right) \times \left( \frac{(L-M + 2z)}{s} + 1 \right) \times K'$-dimensional, where $s$ is the stride (the spatial interval between convolutions centers) and $z$ is the number of 0-valued rows and columns added at the borders of the image, or \emph{zero padding}. %Note that this size must be an integer.
Zero-padding the inputs is very important to control the size after convolution (e.g. to ensure an activation for each location with respect to the input). Convolutional layers are not fully connected: neurons are \emph{shared}, i.e. each filter is applied by sliding it over the whole input, without needing to learn a specific neuron per location. Response (activations) for each filter are then stacked and passed forward. An example is given in Fig.~\ref{fig:conv}(a).

\paragraph{Nonlinear activations}

The neural network community explored the use of many saturating nonlinearities. %, whose outputs tend to fixed values as the magnitude of the inputs grows. The most common ones are the sigmoid and the \emph{tanh}. 
However, such nonlinearities can significantly slow down or even block weight convergence during training, since the gradients tend to zero when inputs magnitudes are large, making null or very small updates (a problem known as the \emph{vanishing gradient}~\cite{xu2105icmlw,he2015cvpr}). For this reason, new (nonsaturating) nonlinearities have been proposed to improve gradient propagation and ultimately convergence and generalization accuracy. The most common activation employed is the Rectified Linear Unit (ReLU), formulated as $\x'_{k'} = \max(0,\x_{k'})$ \cite{nair2010icml}.

Being nonsaturated, ReLU does not suffer from vanishing gradients. It can be computed very efficiently and it naturally sparsifies activations. However, sparsification can also be a drawback, since it can permanently ``kill'' a neuron: if an activation is below zero, it will never be re-activated, since the ReLU will not gate the gradient. To cope with this issue, \cite{maas2013icml} proposed a variant of ReLU called ``Leaky ReLu'' (lReLU), allowing the propagation of gradients also for neurons that would have been deactivated: 
\begin{equation}
\x'_{k'} = \llbracket \x_{k'} \geq 0 \rrbracket \cdot \x_{k'} + \llbracket \x_{k'} < 0 \rrbracket \cdot \tau \,\x_{k'} 
\end{equation}
where $\llbracket \cdot \rrbracket$ is the Iverson brackets, returning 1 if the condition in the brackets is true, 0 otherwise. The scaling $\tau$ is usually a small number allowing small gradients to propagate and coping with dying neurons.

\paragraph{Spatial Pooling}

The spatial pooling layer has the function to summarize the signal spatially (downsamplings), preserving discriminant information. It additionally promotes translation invariance, by pooling over small windows (typically 2$\times$2 or 3$\times$3) into single values. Therefore, pooling layers allow the model to recognize object instances independently on their location. In a classification setting, an image containing an airport will be labeled as ``airport'' independently on where the airport is spatially located within the image. % and this property is due to the spatial pooling of local scores across layers. 
The pooling layer operates on each dimension of the activations independently. Standard pooling strategies are average and the max pooling. The former returns the average of a group of activations in the $P\times P$ window centered on $ij$, denoted as $P_{ij}$ as $\x'_{ij} = \frac{1}{\lvert P_{ij} \rvert} \sum_{a \in P_{ij}} \x_a$, while the second returns the maximum value in $P_{ij}$, as $\x'_{ij} =  \max_{a \in P_{ij}} \x_a$. It has been observed that average pooling might not preform well, since small activations can cancel out larger ones. Max-pooling tends to perform much better since it propagates only information pointing out the presence / absence of some particular feature. However, it might tend to overfit more easily training data. Very strong activations can control the learning of convolutions at lower layers, since backpropagation only makes the gradients flow through the maximum value occurring in each pooling window~\cite{zeiler2013iclr}.

% \begin{figure}[!t]
% \centering
% \begin{tabular}{rl} 
% \raisebox{1mm}{\rotatebox{90}{\small CNN-PC}} & \includegraphics[height=2.5cm]{\figdir net-patchclassifier.png} \\
% \raisebox{1mm}{\rotatebox{90}{\small CNN-SPL}} & \includegraphics[height=2.5cm]{\figdir net-smallfieldpatchpred.png}\\
% \raisebox{1mm}{\rotatebox{90}{\small CNN-FPL}} & \includegraphics[height=2.5cm]{\figdir net-fdeconv-lowres.png} \\
% \end{tabular}
% %\vspace*{-4mm}
% \caption{Schematic of the networks used to perform semantic labeling. \label{fig:fullnet}}
% \vspace*{-4mm}
% \end{figure}

\paragraph{Deconvolutions}

Deconvolution is the transposed convolution operator. A standard convolution outputs a single value per filter and location. This activation is a combination of the input values and the learned filters. The repeated application of convolutions performs a spatially weighted average of the original signals. By combining this with the downsampling nature of max pooling or with the use of convolutions with a stride larger than one pixel, the output activations are downsampled with respect to the original resolution. Typically, downsamplings reduce the spatial activations by a factor of 2.

If one wants to perform an in-network upsampling, activations of the previous layer can be relocated in the upsampled grid and values interpolated by a \emph{deconvolution} operator. This allows to upsample spatially the signals, while still learning channel-wise filters increasing the expressive power of the model. An example of 2$\times$ upsampling is given in Fig.~\ref{fig:conv}(b), employed three times in the upsampling block of our system (blue box in Fig.~\ref{fig:fullnet}). It is important to remark that the deconvolution operator simply corresponds to the backward-pass implementation of the standard convolution (by mapping max-poling activations back, i.e. ``unpooling''). Inversely, the backward-pass of the deconvolution corresponds to a convolution. Therefore, most implementations of CNNs already include such operation. %Deconvolutions are thus learned interpolators / upsampling filters for each feature channel. 
Learned deconvolutions can be followed by nonlinearity layers (e.g. ReLU) in order to learn nonlinear upsampling filters. In our implementation, we employ deconvolutions to compensate the max-poolings, thus mirroring the number of downsampling operations. Differently from other deconvolution layers in CNNs (e.g. the fully convolutional network in \cite{long2015cvpr} or the deconvolutional network in \cite{noh2015iccv}) we only upsample with a 2$\times$ factor at every layer and all the deconvolution filters are learned (we do not fix or initialize any layer to bilinear upsamplings as in \cite{long2015cvpr}). By such small upsamplings we directly obtain geometrically accurate labelings, without the need for unpooling or combining with the activations of lower (higher resolution) layers.% The proposed CNN-FPL implements deconvolutions in the blue box in Fig.~\ref{fig:fullnet}.

% (with downsampling factor of 2 given by max pooling) back to the original input size, by upsampling with a factor 2. By combining downsampling and upsampling, we are able to learn a compression of high-level concepts with a relatively small number of parameters, and then learn an upsampling to the original input patch size, \emph{end-to-end}. 
%While standard convolutions associate activations contained in the neuron to a single output, deconvolutions interpolates inputs to higher resolution outputs by means of learned filters. Example of a deconvolution compared to a convolution operation is illustrated in Fig.~\ref{fig:conv}(a)-(b).

\paragraph{Classification layer and loss}

The topmost layer (red boxes in Fig.~\ref{fig:fullnet}) is composed by a classifier, whose loss is differentiable. We use the commonly employed Multinomial Logistic Regression, whose scores (class-conditional probabilities) are given by the \emph{softmax} function:
\begin{equation}
p(y_i|\x_i) = \frac{\exp(\x_i)}{\sum_{c=1}^C \exp(\x_{ic})}
\end{equation}
for $C$ classes. Inputs $\x_i$ are a $C$-dimensional vector representing unnormalized scores for the location $i$, as given by the penultimate layer (the one before the loss function). The filters of this penultimate layer can be interpreted as the weight vector of the classifier. The classification loss (cross-entropy) is:
\begin{equation}
L(y_i) = - \sum_{c=1}^C \llbracket y_{p(y_{ic})} = c \rrbracket  \log\left(p(y_{ic}|\x_i)\right).
\end{equation}
This loss function trains the network by forcing it to put all the mass on the correct labeling. In our dense labeling setting, the loss is not computed over a single prediction as for the CNN-PC, but over a the grid of spatial predictions. During training, the actual value of $L(y)$ corresponds to the average per-patch loss, over the training batch. The loss for each patch is again the average per-location loss (at each predicted pixel). 
%The loss for the batch of training samples is the average loss for image patches.% The latter is obtained simply by summing over all the patch locations and dividing.
During inference, the predicted label for location $i$ will be given by $y_i^* = \argmax_c p(y_{ic}|\x_{i})$.

% \renewcommand{\tabcolsep}{5pt}
% \begin{figure}
% \centering
% \begin{tabular}{c|c}
% $\overbrace{\includegraphics[height=2.5cm]{\figdir downsamplingblock.png}}^{\text{Downsampling block}}$ &
% $\overbrace{\includegraphics[height=2.5cm]{\figdir upsamplingblock.png}}^{\text{Upsampling block}}$ \\
% \end{tabular}
% \caption{Building blocks of the proposed CNN. 1. Input to layer $l$ from layer $l-1$; 2 Batch normalization; 3. ReLU; 4. Max pooling; 5. Dropout; 6. Input to $l$, from $l-1$; 7. Batch normalization; 8. ReLU; 9. Dropout. \label{fig:convblocks}}
% \vspace*{-4mm}
% \end{figure}

\subsection{Mitigating overfitting}

In this paper, we employ these five layers to build the CNNs illustrated in  Fig.~\ref{fig:fullnet}. Although the datasets are large and the number of parameters is not prohibitive, training of such models might be difficult. The information content in such data is heavily redundant and most semantic classes are characterized by relatively uniform, while most of the variations in the data lie in the spatial arrangement of such classes. %Therefore, when looking at the data through fixed sized windows, we often see the same elements only arranged differently. 
Predicting a single label per patch is suboptimal, since the model is not explicitly taking advantage of these regularities while learning, but it is only trained to predict their presence in the (center of) the patch. %In the semantic labeling setting, 
To learn such spatial arrangement we might need models with larger capacity, since deconvolutions are needed after the bottleneck layer. However, this corresponds to optimizing over a larger number of parameters. %and therefore increasing the risk of overfitting. 
In this section, we will review a series of strategies to cope with overfitting.

\setcounter{paragraph}{0}
\paragraph{Dropout}
This technique has been proposed to avoid co-adaptation of neurons during training \cite{srivastava2014jmlr}. Co-adaptation would result in filters in the same layer which are inter-dependent one to each other. Therefore, such network would be harder to train and ultimately fitting too tightly training data, without any good generalization guarantee. Dropout draws from ensemble learning theory: randomly ``turning off'' a given percentage of neurons (dropout rate hyperparameter) and their connections, corresponds to train a different, less correlated, model at every epoch. At inference time, an approximate ensemble method is obtained activating all the connections. In practice, dropout could slow down training, but benefits largely surpass drawbacks, in particular when the number of parameters to learn is large (e.g. fully connected layers).

%In our network, we apply dropout at every convolutional and deconvolutional layers. At each iteration, dropout switches off neurons randomly with a retention probability of 0.5.

\paragraph{Batch Normalization}

Batch normalization \cite{ioffe2015icml} aims at speeding up training by allowing the use of larger learning rates and mini-batches. To do so, it learns the normalization for each batch so that the activations passed to the next layer follow a normal distribution with $\mathcal{N}(0,1)$. This might seem trivial, but it avoids problems related to the drift of activation distributions during training and makes the whole system less sensible to layer initialization. Moreover, by keeping values normalized at each layer, difference in the randomly selected mini-batches and the activations they generate should have less influence on the weight updates across iterations. Each layer can now focus on the general improvement rather than learning to adapt to the previous updates. We included this layer right after every convolutional and deconvolutional layer.

\paragraph{Weight decay}

Weight decay is an $\ell_2$ regularizer adding a penalty term to weight updates during backpropagation. It is applied only to convolutional filter weights (and not to the biases) and favors smooth convolutional filters. The weight decay hyperparameter controls the penalization. 

\paragraph{Data Augmentation}

%CNN need to learn a large amount of parameters and consequently they require large amounts of training data. It has been shown that, if the amount of labeled data is insufficient, learning relevant filters is hindered and the model performs as well as with random filters~\cite{saxe2011icml}. 
A commonly used strategy to further increase the size of the training set, is to perform \emph{data augmentation}. It consists in creating new synthetic training examples from those already available, by applying label-preserving (random) transformations. This step ensures that the model sees different possible aspects of the data in different batches, improving generalization error by i) increasing the number of labeled samples to learn from and ii) regularizing the model \cite{bishop1995neucomp} and iii) reduce potential correlation between patches in the batch. To train the CNNs, we first create a \emph{super-batch} by sampling a given number of training patches. From this super-batch, we then randomly sample the mini-batch used to effectively train the CNNs. We apply transformations at both levels. We describe adopted augmentation strategies below:

%\setlist[description]{labelindent=\parindent,style=multiline,leftmargin=0.5cm}
%\begin{description}
{\it - Random sampling.} Semantic classes are unevenly distributed spatially and their frequency highly varying. For instance, the class ``road'' is ubiquitous, while the class ``car'' is localized (appearing mostly on roads) and rare (in the Vaihingen dataset described in Section~\ref{sec:data}, 1.2\% of the training ground truth represents ``cars'', while 27.94\% ``roads''). Consequently, we generally sample the training patches randomly in space (among the training images) and uniformly with respect to class frequencies. To control this process, we simply account for the class corresponding to the label of the central pixel of the patch. % This does not augment the dataset directly, but it increases variance in the super-batch. %As the probability of sampling exactly the same locations is very low, the batches sampled this way will always be different. %The random selection of the training batch further includes randomness, however, since classes in the super-batch are uniformly represented, this does not pose a problem.

{\it - Random transformations.} %A significant number of image patches are very similar, but not identical. As the dataset grows, little information is gained if drastically new objects do not appear repeatedly. %Even in this latter case, a single new object will more likely be considered an outlier rather than a regularity in the data. 
To slightly vary the spatial organization of the patches and to enforce learning invariances of interest, we randomly rotations at random angles and flippings. Random rotations are applied on each training image before sampling the super-batch (as explained later in Sec.~\ref{sec:ourmodel}, the super-batch will be resampled at regular intervals). %Secondly, from each training tile we crop a randomly sized patch (from a bounded uniform distribution, whose bounds are hyperparameters) and resize it to the input size required by our network architecture. 
Random flippings are applied when selecting the mini-batch during training, independently to rotations.

{\it - Noise injection.} The last class-preserving random transformation consists in jittering, i.e. injecting small random additive noise to each patch in the mini-batch. Jittering is important since it forces the model to learn spectrally smooth decision rules, by reducing correlation across similar patches in the mini-batch and making the input distribution denser. %Since dealing with aerial images requires accounting for strongly textured areas (including similar textures for different classes or different textures for the same class), 

The noise is sampled from the Normal distribution $\mathcal{N}(0,0.01)$. Note that the input data is scaled in $[0,1]$.
%\end{description}

%%% Local Variables: 
%%% mode: latex
%%% TeX-master: "main-volpi-cnn-deconv-journal"
%%% End:
% !TEX root = main-volpi-cnn-deconv-journal.tex
\section{CNN architectures considered in the paper}\label{sec:ourmodel}

%In this paper, we adopt an architecture based on a downsampling-then-upsampling logic to perform semantic labeling of images by fully preserving the input data resolution (CNN-FPL). We compare it against a common architecture performing patch classification (CNN-PC) and to a second less standard setting predicting small patches of labels (CNN-SPL). 
%This section presents the three adopted architectures (CNN-PC, CNN-SPL, CNN-FPL) in detail.

\subsection{Common strategies}

Besides architecture-specific training strategies (detailed below for each CNN), we define some general rules to train each architecture, in particular with respect to sampling training patches. To train each model, we first sample the super-batch set composed by $N^\text{trp}$ training patches. We define every training \emph{epoch} to be composed of 500 passes over the super-batch set. The super-batches (at both training and testing) are centered around the mean values of the training set.  %We explicitly train it on a single scale, thus providing to the network fixed sized inputs, as the resolution and scales across images is constant. %To ensure sufficient diversity, but also to allow the network to learn recurrent structures, w
We set its size as  $N^\text{trp} = N_b \cdot 500$, uniformly sampled across the training tiles. $N_b$ denotes the mini-batch size, which depends on the architecture. Each patch is sampled randomly in the spatial domain. %We explicitly train it on a single scale, thus providing to the network fixed sized inputs, as the resolution and scales across images is constant.%, with a side size selected from $[61,63,65,67,69]$. Then, the patch is resized to be 65 $\times$ 65, as input for all the architectures. 
We resample the super-batch every 20 epochs for the Vaihingen and 5 for Potsdam. 
For all the models, input patches are 65$\times$65 pixels in size.

For all the convolution / deconvolution filters described in the following subsections, weights are initialized from $\sqrt{\frac{2}{M^2\cdot K'}}\mathcal{N}(0,1)$ (improved Xavier initialization) and are all applied with a stride = 1.

%\blue{For all the presented models, we employ 64 neurons at the first layer (7$\times$7 filters), 64 at the second (5$\times$5), 128 at the third and finally (5$\times$5) 256 in the fourth layer. For the CNN-PC, the fourth layer includes max-pooling and relu, and then implements a fully connected layer to map activations to single class scores, per patch.
%For the CNN-SPL, the last max-pooling is removed and the fully connected layer of the CNN-PC is replaced by $1 \times 1$ convolutions, so that the output is now a patch of labels of the same size as the bottleneck layer.
%For the CNN-FPL model, the $1\times 1$ convolutions are replaced by 3$\times$3, 512-dimensional deconvolution layers, doubling the size of the activation of the previous layer. We employ 3 deconvolution layer to compensate 3 levels of downsampling (max-poling). The $65 \times 65 \times 512$ activations are mapped to class scores through $1 \times 1 \times 6$ convolutions, to obtain a $65 \times 65 \times 6$ score map. 
%All the convolution / deconvolution are followed by a batch-normalization layer, except for the layers before the softmax. Leaky ReLUs with a leak parameter of 0.1 are used right after the batch normalization, and right before 3$\times$3, stride 2 max-pooling operations. 
%All the convolution / deconvoltion filter weights are initialized from $\sqrt{\frac{2}{M^2\cdot K'}}\mathcal{N}(0,1)$ (improved Xavier initialization) and are applied with a stride 1. Examples of convolution / deconvolution blocks are given in Fig.~\ref{fig:convblocks}.}

We employ backpropagation with stochastic gradient descent (SGD) with momentum~\cite{krizhevsky2012nips}. We fix the momentum multiplier to the standard value of 0.9. To monitor the validation error during training we predict, at each epoch, $N_b \cdot 100$ validation patches from the validation images, and compute the error. We also sample validation patches uniformly across classes. Note that validation patches are never used in the training process. 

% Moreover, as we sample patches uniformly across classes, the validation error is not representative of the final segmentation accuracy. %Finally, as the class ``clutter'' / ``background'' is very different in appearance (it contains different classes) and very small with respect to the total number of pixels, we propose a specific strategy to deal with such cases. We train the networks for 300 epochs without showing any example of ``clutter''. Then, from epoch 301 on, we start showing to the model examples of this class with a small learning rate. We found that this strategy allows a better training convergence and ultimately allows to learn the general appearance of most common ``clutter'' objects.

\subsection{Architecture 1: patch classification (CNN-PC)}

\renewcommand{\tabcolsep}{1pt}

%\begin{figure}[!t]
%\centering
%\begin{tabular}{rlrlrl} 
%\raisebox{1mm}{\rotatebox{90}{CNN-PC}} 
%& \includegraphics[height=2.2cm,angle=90]{\figdir net-patchclassifier.png} & %\raisebox{1mm}{\rotatebox{90}{CNN-SPL}}
%& \includegraphics[height=2.2cm,angle=90]{\figdir net-smallfieldpatchpred.png} & %\raisebox{1mm}{\rotatebox{90}{CNN-FPL}}
%& \includegraphics[height=2.2cm,angle=90]{\figdir net-fdeconv-lowres.png} \\
%\end{tabular}
%\vspace*{-4mm}
%\caption{Schematic of the networks used to perform semantic labeling. \label{fig:fullnet}}
%\vspace*{-4mm}
%\end{figure}

\noindent{\textit{Structure:}} This architecture is a standard patch-classification system, i.e. a model that takes as input a patch and predict the label of the patch (the label of the central pixel of the patch). We employ 64 neurons at the first layer (7$\times$7 filters), 64 at the second (5$\times$5), 128 at the third and finally (5$\times$5) 256 in the fourth layer. The fourth layer includes max-pooling and relu, and then implements a fully connected layer to map activations to single class scores, per patch. A schematic is illustrated in Fig.~\ref{fig:fullnet}. Leaky ReLUs with a leak parameter of 0.1 are used right after the batch normalization layers, and right before 3$\times$3, stride 2 max-pooling operations. Dropout with a 50\% rate is applied at each layer. No batch normalization or dropout is applied before the softmax.

\noindent{\textit{Training:}}  Learning rates for SGD start at $10^{-3}$ then reduced by 0.5 every 100 epochs, until 300 epochs are completed. Then, it's set to $10^{-5}$ for the last 100 epochs. The weight decay is fixed to 0.01. The network is trained for a total of 400 epochs with mini-batches of size of 128. 

\noindent{\textit{Inference:}} At inference time, we predict a single label per patch. As discussed in the Introduction, patch-based strategies such as CNN-PC do not offer the ability of predicting directly structured spatial arrangements of pixels, since inference is done by spatially independent predictions. To provide a dense pixel map, different strategies can be adopted. The simplest na\"ive approach is to decompose the image into a series of overlapping patches of size corresponding to the one of the CNN input and predict the class for each region independently. In this paper, we adopt this prediction strategy mainly to avoid any potential bias introduced by other techniques. However, as we will detail in the Results section (Section~\ref{sec:resu}) we performed inference with a stride larger than 1 and then we interpolated results back to the original resolution with a negligible loss in accuracy.

%Since want the prediction to have the same resolution as the test image, %each  iterates over pixels of the prediction images, which is extremely costly .%expensive computationally and can not make use of the principle of shared convolutions over the whole image extent. 
%We observed that predicting with stride = 1 or stride = 2 and then upsampling class-conditional probabilities only reduces performances by less than 1\% in OA, while reducing by 4 inference time.

An alternative way to obtain dense predictions is to employ the Dense Neural Pattern strategy, presented in \cite{wang2015tpami} and used in \cite{paisitkriangkrai2015cvprw}. %Rather than producing a dense semantic labeling or a dense prediction of class-conditional scores, this approach extracts features for each location, which need to be subsequently processed by a classifier. At each convolutional layer (or at a specific convolutional layer), this approach extracts activations to be used as features. 
A similar approach is presented in \cite{hariharan2015cvpr}, where stacked activations are defined as ``hypercolumns''. 

%Since all of these approaches perform independent pixel/location-wise predictions, it is common to employ post-processing by Markov/Conditional random fields, which are not applied after the CNN-PC (nor after any model implemented in this paper).

\subsection{Architecture 2: subpatch labeling (CNN-SPL)}

\noindent{\textit{Structure:}} The second CNN we train for comparison predicts the labeling of a sub-patch of size 9$\times$9 around the 65$\times$65 input patch center. This network corresponds exactly to the CNN-PC, but the last max-pooling is removed and the fully connected layer of the CNN-PC is replaced by $1 \times 1$ convolutions, so that the output is now a patch of labels of the same size as the bottleneck layer. Leaky ReLUs with a leak parameter of 0.1 are used right after the batch normalization layers, and right before 3$\times$3, stride 2 max-pooling operations. Dropout with a 50\% rate is applied at each layer. No batch normalization or dropout is applied after the $1\times 1$ convolutions.

%Instead, we utilize a layer composed by $1\times 1 \times 6$ convolutions with no zero-padding and stride 1, mapping the $9 \times 9 \times 256$ activations into a $9 \times 9 \times 6$ class-conditional scores. This last layer, as opposed as the fully connected one of the CNN-PC network, learns to map single locations to class scores rather than combining the activations into a single class score.

\noindent{\textit{Training:}} CNN-SPL is trained in the same way as CNN-PC described above. We use the weights of the CNN-PC to warm start the training procedure. For this reason, learning rates for this network architecture are 10 times smaller than the ones employed for the CNN-PC (but the same number of epochs), as we basically only fine tune the system to the new output. Mini-batches are composed by 128 examples.

\noindent{\textit{Inference:}} %At inference time, CNN-SPL predicts labels for a $9 \times 9$ patch centered on the input $65 \times 65$ input patch. It does so by cutting the fully connected layer from the standard CNN-PC and replacing it with $1\times 1$ convolutions, trained to locally predict labels in a $9\times 9$ window around the center pixel. 
Because of the special structure in predicting patches of labels, a feedforward pass of the whole image would produce a prediction with a resolution roughly 3 times higher compared to the CNN-PC. We upsample the outputs of the last layer (softmax scores) using bilinear interpolation to match the size of the original image tile.

\subsection{The proposed architecture: full patch labeling by learned upsampling (CNN-FPL)}

\noindent{\textit{Structure:}} This architecture is composed by two main parts: a downsampling (green block of Fig.~\ref{fig:fullnet}) and an upsampling block (blue area of Fig.~\ref{fig:fullnet}). The downsampling part corresponds exactly to the CNN-SPL, but the $1\times 1$ convolutions are replaced by 3$\times$3, 512-dimensional deconvolution layers, doubling the size of the activation of the previous layer. We employ 3 deconvolution layer to compensate 3 levels of downsampling (max-poling). The $65 \times 65 \times 512$ activations are mapped to class scores through $1 \times 1 \times 6$ convolutions, to obtain a $65 \times 65 \times 6$ score map. %Main blocks of this architecture are detailed in Fig.~\ref{fig:convblocks}. % The only difference between the blocks resides in the use of deconvolutions at the upsampling step and, instead of applying max-pooling reducing the spatial dimensions by half, an upsampling factor of 2$\times$ is applied to the input. The upsampling layers are directly stacked on the activations of the last convolutional layers.%, the 9$\times$9 layer coming from the previous CNN-SPL, without the last 1$\times$1 convolutions.  
As for previous architectures, batch normalization is applied after every convolution / deconvolution and  dropout is applied with a 50\% rate. No dropout nor batch normalization is applied after the 1$\times$1 convolutions. Leaky ReLUs are employed with a leak factor of 0.1 and max-pooling layers in the downsampling part are exactly as for the other architectures.  

We argue that the bottleneck structure is beneficial, since it forces the network to learn a rough spatial representation for the classes present in the patch (after the downsampling block, the original $65 \times 65$ patch is reduced to a 9$\times$9 map of activations), before learning upsampling of these representations back to the input resolution. Note that the embeddings of the CNN-SPL and CNN-FPL architectures at the 9$\times$9 activations layer are completely different: in the former, the network learns to predict a small patch corresponding to the central area of the input region. Thus, the concepts embedded in the penultimate layer directly correspond to the arrangement of pixels in the original resolution image. In the proposed CNN-FPL network, the concepts in the bottleneck correspond to a degraded resolution of the entire input patch, pointing to main features in a coarser geometry. The upsamplings are learned so that the deconvolutions reconstruct learned spatial and geometrical arrangements of activations at larger scale but acting locally (3$\times$ 3). %To do so, the upsampling block learns a series of small filters (3 $\times$ 3) to upsample by a factor 2 the inputs to the deconvolutional layers. These layers also learn new nonlinear representations by including lReLU layers, as illustrated in Fig.~\ref{fig:convblocks}. 
This strategy allows to interpret the output of the CNN-FPL as structured, since every predicted label is learned to be interdependent with its neighbors, conditioned on the receptive field of the previous layers.% The upsampling deconvolution layers %are all initialized as bilinear upsampling layers, as suggested in \cite{long2015cvpr}, and count all 512-dimensional filters. After the last upsampling layer, the $65 \times 65 \times 512$ activations are mapped to class scores through $1 \times 1 \times 6$ convolutions, to obtain a $65 \times 65 \times 6$ score map. 

\noindent{\textit{Training:}}  CNN-FPL is trained in the same way as architectures described above. However, since it involves the upsampling layers, the number of learnable parameters is significantly higher than for the two previous networks. We warm start the system by employing weights from the CNN-PC, for the common downsampling blocks. The learning rates are initialized at $10^{-3}$ and kept uniform for 100 epochs, then decreased to $5\cdot10^{-4}$ and $10^{-4}$ for 100 epochs each. A last 300 epochs with $10^{-5}$ are performed to fine tune weights. Weight decay is the same as for the other two models, but we use a smaller mini-batch size of 32 examples. Since during training the CNN-FPL sees less patches, we trained it for an additional 100 epochs.

\noindent{\textit{Inference:}} At inference time, the proposed CNN-FPL approach produces a segmentation of same size as the input patch by a single forward pass of the image. This way, the three main drawbacks of the aforementioned systems are circumnvented: first, the whole image can be feedforwarded to the trained CNN and we can directly obtain a dense labeling (stride is equal to 1 by construction) or a dense class-conditional score map; second, we do not need a second step of prediction upsampling; and third; the prediction performed this way is locally structured, as the CNN is able to exploit both color and semantic correlations represented in the input patch, without the need of a subsequent structured output post-processing.

\subsection{On the choice of the architecture and general setup}

We tested different architectural setups by evaluating the accuracy on the validation set. The final CNN-FPL network architecture is driven by mainly 3 factors: Firstly, the number of downsampling layers has been fixed to 3, after testing also 2 and 4 max-poolings layers (and consequently the total number to 6 plus one 1$\times$1 score map layers). The former made the network too shallow, while the latter too deep to optimize. The solution in the middle offered the best trade-off. Secondly, we tested patch sizes large enough to include some spatial context, in particular local co-occurrences and allowing filters to learn recurrent structures. We trained for few iterations network with varying input patch sizes in $\{25,45,55,65,85\}$ and observed that the size of 65 pixels side offered a good compromise between accuracy and memory requirements. The size of the input patch also directly influences the spatial extent at the bottleneck layer: 65$\times$65 are reduced to 9$\times$9 at the bottleneck, whose extent contains enough spatial information to allow direct and geometrically accurate upsamplings. Larger bottleneck provide similar accuracy, but requiring more memory during training. Thirdly, the number of neurons has been chosen by training again for few iterations a series of networks with varying number of neurons at the first layer, growing in power of two (starting from 16). We follow the roughly the rule that a layer has double the number of channels of its predecessor. Since the number of channels directly influences overfitting, it directly controls the number of learnable parameters, we did not selected more than 512 channels. Architectures of the competing CNN models are all based on the CNN-FPL architecture, as described above (by removing the deconvolutional part we obtain the CNN-SPS network and by including a fully-connected layer on top of it we obtain CNN-PC). This allows us to directly compare the effects of the architectures. Finally, we selected the batch size experimentally on the validation set and it has been fixed to 128 for CNN-PC and CNN-SPL, while to 32 for the CNN-FPL. The smaller batches cause slower convergence but ease the problem of overfitting.

All the presented results are computed on a desktop machine with an Intel Xeon E3-1200 (QuadCore), 32Gb RAM and an Nvidia GeForce GTX Titan X (12Gb RAM). We build the tested system using the DAGNN wrapper around CNN libraries provided by MatConvNet\footnote{\url{http://www.vlfeat.org/matconvnet/}} version 1.0 beta 20.

\section{Data and experimental setup}\label{sec:resu}

\subsection{Dataset Description}\label{sec:data}

We evaluate the proposed system on the Vaihingen and Potsdam datasets provided in the framework of the 2D semantic labeling contest organised by the ISPRS Commission III\footnote{http://www2.isprs.org/commissions/comm3/wg4/semantic-labeling.html}. The Vaihingen data is composed by a total of 33 image tiles (average size of 2494 $\times$ 2064), 16 of which are fully annotated. The remaining tiles compose the test set. Each image is composed by near infrared (NIR), red (R) and green (G) channels with a spatial resolution of 9cm. We also dispose of a digital surface model (DSM) coregistered to the image data, which has been normalized ({\it n}DSM) and redistributed in \cite{gerke2015techrepo}. In this work we jointly use spectral information (NIR-R-G) and the {\it n}DSM to train the network. An example of image tile is given in the first three panels of Fig.~\ref{fig:fullpred34}. We use 11 out of the 16 annotated images to train the networks and the remaining 5 to validate training and to evaluate the segmentation generalization accuracy (ID 11, 15, 28, 30, 34). The input of each network corresponds to stacked NIR-R-G and {\it n}DSM, for a total of 4 input dimensions. All the data available are rescaled into the $[0,1]$ interval. When comparing to state-of-the-art on this dataset, we refer to results published on the challenge website\footnote{http://www2.isprs.org/vaihingen-2d-semantic-labeling-contest.html}.

The Potsdam 2D semantic labeling challenge dataset\footnote{http://www2.isprs.org/commissions/comm3/wg4/2d-sem-label-potsdam.html} features 38 tiles of size 6000$\times$6000 pixels, with a spatial resolution of 5cm. From the available patches, 24 are densely annotated, with same classes as for the Vaihingen dataset. This dataset offers NIR-R-G-B channels toghether with DSM and normalized DSM. We employ all spectral channels plus the DSM as input for our networks (5 dimensions). In our setting, we use the tiles 02\_12, 03\_12, 04\_12, 05\_12, 06\_12, 07\_12 for validation, and the remaining 18 for training. As for the Vaihingen case study, we employ as input all the spectra information and the {\it n}DSM, rescaled in $[0,1]$. %We employed the ``ourapproach'' {\it n}DSM when possible, otherwise we sampled from the ``lastools'' one.}

\subsection{Competing method}

Besides the CNN-PC and CNN-SPL architectures, acting as strong baselines, we also compare CNN-FPL to a modern baseline implementing standard superpixel-based labeling with hand-crafted features. For each image, we stack raw color information, Normalized Difference Vegetation Index, Normalized Difference Water Index using G and NIR channels and normalized DSM. For each feature we extract 4 mathematical morphology operators (opening, closing and their reconstructions) as well as a texture statistic (entropy). Each filter is operated in 3 window sizes (7, 11, 15 pixels). We then extract regions using the graph-based superpixelization of  \cite{felzenszwalb2004ijcv} on the channels of the raw data. We summarize the distribution of appearance descriptors in each superpixel by extracting minimum, maximum, average and standard deviation of the values. The final feature set is composed of 384 features. We then train a random forest classifier on this set by training 500 trees with 100 random feature tests per split, and retaining the one with optimal Gini diversity criterion. Each tree is trained with the full training set composed of 5000 randomly selected example \emph{per class}, over the 11 training images. A leaf in the tree is obtained when only 2 examples remain at the node. We refer to this approach as superpixels with multiscale features (SP-MSF). We use the same strategy on both datasets.

\subsection{Evaluation Metrics}

For both datasets, we present figures of merit obtained on the corresponding validation sets. We employ four different evaluation strategies. The first -- {\bf full} -- is a standard full spatial evaluation with all classes present in the ground truth. The second strategy -- {\bf no bk} -- excludes from the evaluation the class ``background'', but preserves the full spatial extent of the ground truth. The last two -- {\bf er full} and {\bf er no bk} -- estimate figures of merit as in the first two strategies, but after eroding the edges of each class in the ground truth with a 3 pixel circular structuring element, so that evaluation is tolerant to small errors on object edges. We report overall accuracy (OA), Kappa (K), average class accuracy (AA), class-averaged F1 score (F1). The OA and K are global measures of segmentation accuracy, the former providing information about the rate of correctly classified pixels and the second compensating for random chance in assignment. Both are biased towards large classes, meaning that contributions of small classes are canceled out by those of larger classes. Both AA and F1 are class-specific and therefore independent from class-size. The former provides the average (per-class) ratio of correctly classified samples (it is therefore insensitive to the size of the ground truth classes), while the latter is the geometric mean between precision (user's accuracy) and recall (producer's accuracy). More details can be found in \cite{congalton2008book}.
 
%\newpage

\section{Results}

\subsection{Vaihingen dataset results}

\subsubsection{Numerical results}

\begin{table}
\renewcommand{\tabcolsep}{3pt}
\caption{Numerical results for the Vaihingen validation set. %We report figures of merit accounting for the background class ({\bf full}) and without it ({\bf no bk}), as well as on by employing the ground truths with eroded edges, on both background ({\bf er full}) and no-background cases ({\bf er no bk}).
\label{tab:validationresults}}
\centering
\begin{tabular}{c|c|cccc}
\toprule
  & {\bf Model}   & {\bf OA}    & {\bf K}     & {\bf AA}    & {\bf F1}    \\\hline 
  & SP-MSF  & 79.79 & 73.28 & 65.80 & 66.20 \\ 
  & CNN-PC  & 82.62 & 76.97 & 62.10 & 64.19 \\
  & CNN-SPL & 83.04 & 77.50 & 64.05 & 67.89 \\
 \raisebox{1mm}{\multirow{-3}{*}{\rotatebox{90}{\bf full}}}   
  & CNN-FPL & {\bf 83.79} & {\bf 78.52} & {\bf 69.12} & {\bf 73.03} \\
\midrule
 % & model  & OA    & K     & AA    & F1    & t [s]  \\\hline
  & SP-MSF  & 79.86 & 73.36 & 74.24 & 74.40 \\ 
  & CNN-PC  & 82.68 & 77.04 & 74.47 & 76.95 \\
  & CNN-SPL & 83.08 & 77.55 & 70.76 & 72.89 \\
 \raisebox{1mm}{\multirow{-3}{*}{\rotatebox{90}{\bf no bk}}}   
  & CNN-FPL & {\bf 83.83} & {\bf 78.57} & {\bf 76.45} & {\bf 78.76} \\
\midrule
 % & m & OA    & K     & AA    & F1    & t [s]  \\
  & SP-MSF  & 83.64 & 78.32 & 70.65 & 69.92 \\ 
  & CNN-PC  & 86.67 & 82.29 & 65.58 & 68.01 \\
  & CNN-SPL & 87.24 & 83.03 & 69.17 & 73.47 \\
 \raisebox{1mm}{\multirow{-3}{*}{\rotatebox{90}{\bf er full}}}
  & CNN-FPL & {\bf 87.80} & {\bf 83.79} & {\bf 74.94} & {\bf 78.60} \\
\midrule
 % & m & OA    & K     & AA    & F1    & t [s]  \\
  & SP-MSF  & 83.71 & 78.39 & 78.44 & 77.64 \\ 
  & CNN-PC  & 86.73 & 82.36 & 78.66 & 81.57 \\
  & CNN-SPL & 87.28 & 83.07 & 75.03 & 77.78 \\
 \raisebox{2mm}{\multirow{-3}{*}{\rotatebox{90}{\bf er no bk}}}   
  & CNN-FPL & {\bf 87.83} & {\bf 83.83} & {\bf 81.35} & {\bf 83.58} \\
\bottomrule
\end{tabular}
\vspace*{-4mm}
\end{table}

Table~\ref{tab:validationresults} presents results for all the tested CNN, and the competing method relying on superpixels. In all the evaluation settings, the CNN-FPL is the most accurate under different accuracy metrics. By removing the ``clutter'' class, the standard CNN-PC baselines gains 11-12 points in AA and F1 scores, indicating that this class was mostly missed. The strategies trained on patch of labels are more robust in this sense, gaining in the range of 4-5 points. By evaluating on eroded boundary ground truths, we observe a similar behavior, but with significantly higher accuracies. This indicates that in all situations the boundaries are often blurred within the 3 pixel erosion radius.  The CNN-SPL shows performances in between CNN-PC and CNN-FPL. It is able to describe the class ``background'', but evaluations without this class bring it at performances lower than the ones of the CNN-PC. %However, recall that this approach performs a single forward pass at inference, and the low resolution maps is upsampled to match the original image size, while CNN-PC classifies each overlapping patch independently. 
All the CNN-based models outperform the baseline employing dense appearance descriptors and superpixel regions. The proposed CNN-FPL results in being the most accurate numerically. We argue that CNN-FPL makes better use of the training data by learning class relationships and co-occurrences, naturally. %We also assume that the co-occurrences are expressed in the learned nonlinear upsampling layers.

CNN-FPL is not only the most accurate method, but it is also extremely fast. As illustrated in Tab.~\ref{tab:validationinferencetime} it only takes 31s (6.2 s/image) to perform inference on the 5 test images (average size of 2563 $\times$ 1810 pixels) on the GPU. The na\"ive inference used for the CNN-PC, i.e. predicting the label of batches of patches independently then rearranging them spatially, requires 13841s (2768 s/image) on the GPU. However, in the rest of the experiments we performed inference with a stride of 2 and then upsampling the probabilities scores with bilinear interpolation. This way, the time can be roughly reduced by two (one prediction per 2$\times$2 pixel grid, instead of 4) by losing very little accuracy ($<$ 1\% OA). Note that there exist different strategies aiming at efficiently speeding up prediction time for CNN. In this comparison, we mainly wanted to point out that learning full patch segmentations results \emph{naturally} in a more accurate and dense prediction. CNN-SPL is the fastest at predicting maps, since it only needs to evaluate few downsampling layers, only requiring 9s to predict maps of the whole validation set (1.8 s/image).

 %Patch-based predictions for CNN-PC can be easily parallelized on GPU to increase efficiency. 
%In our tests, inference for CNN-SPL could not be naturally transferred on the GPU, since images do not fit in the memory. However, inference time on the CPU is still very appealing with 24 s/image. Note that CNN-FPL inference on the CPU is roughly 200 s, while two orders of magnitude larger for CNN-PC.

\begin{table}
\caption{Average inference time per image (on validation set) for the CNN models. \label{tab:validationinferencetime}}
 \centering
\begin{tabular}{c|ccc}
\toprule
        & {\bf CNN-PC} & {\bf CNN-SPL} & {\bf CNN-FPL} \\\hline
 {\bf time (s/image)}  & 2768 (1360.6$^\dagger$)   & 1.8 & 6.2 \\ \hline
\multicolumn{4}{l}{$\dagger$ Inference time with stride 2} \\
\bottomrule
\end{tabular}
\vspace{-4mm}
\end{table}

\subsubsection{Qualitative Results}

\begin{figure*}
\centering
\begin{tabular}{rrccc|cccc}
& {\bf ID} & {\bf Image} & {\bf nDSM} & {\bf GT} & {\bf SP-MSF} & {\bf CNN-PC} & {\bf CNN-SPL} & {\bf CNN-FPL} \\ \hline
& \raisebox{8mm}{\bf 1} & 
\includegraphics[width=.13\textwidth]{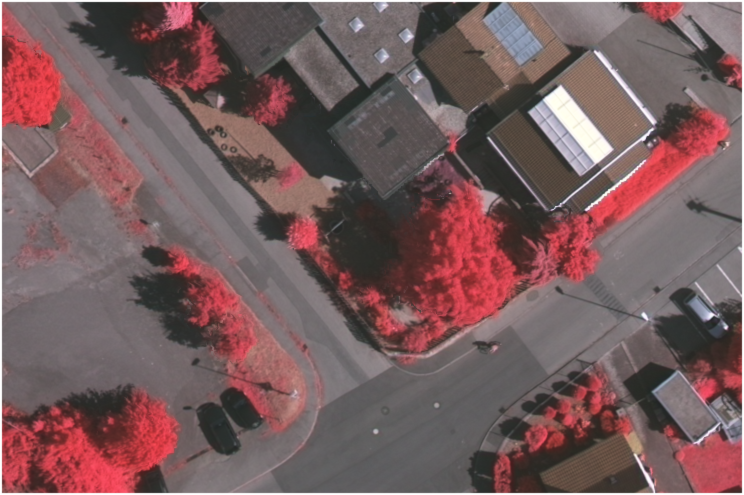} & 
\includegraphics[width=.13\textwidth]{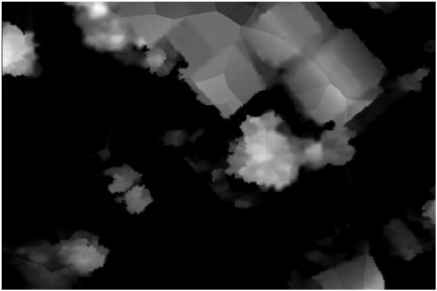} & \includegraphics[width=.13\textwidth]{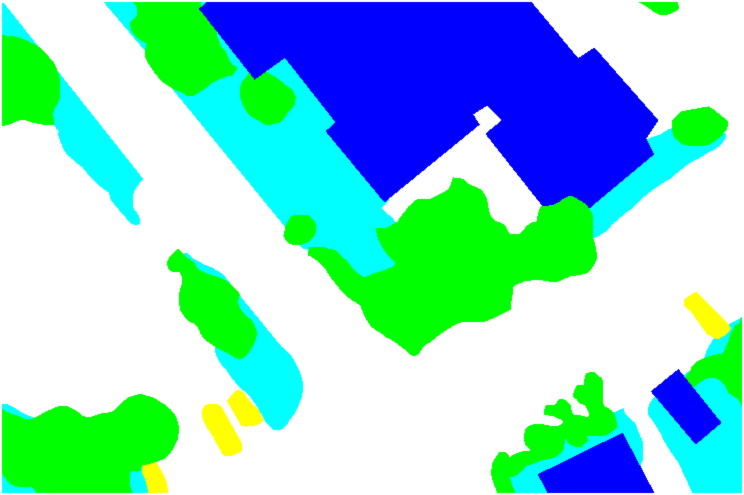} & \includegraphics[width=.13\textwidth]{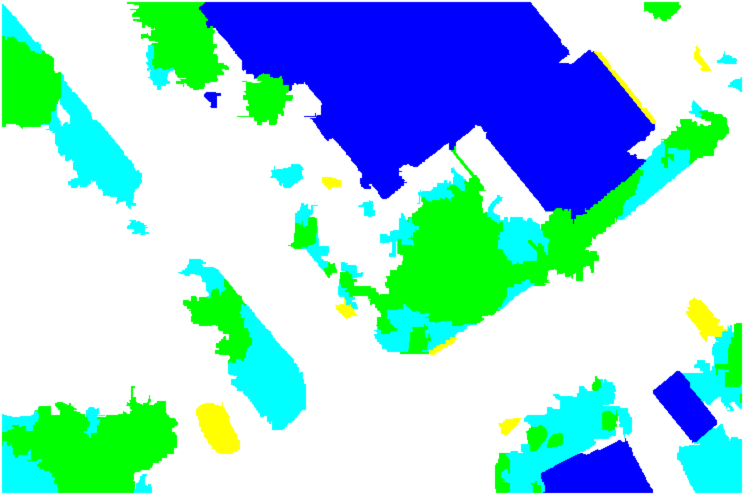} & \includegraphics[width=.13\textwidth]{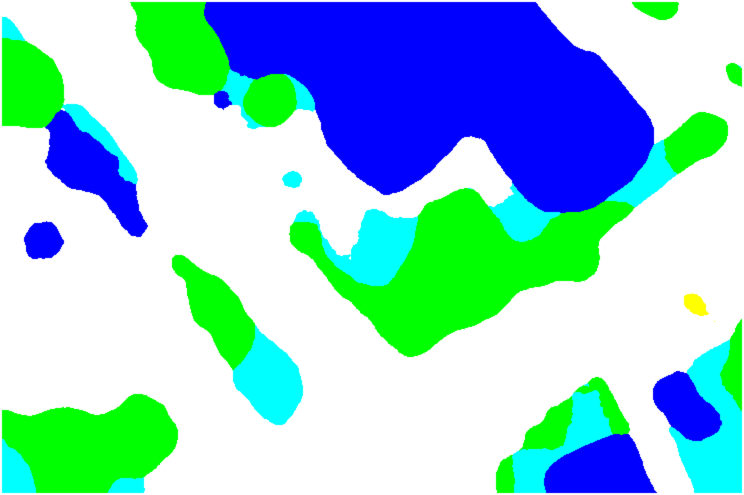} & \includegraphics[width=.13\textwidth]{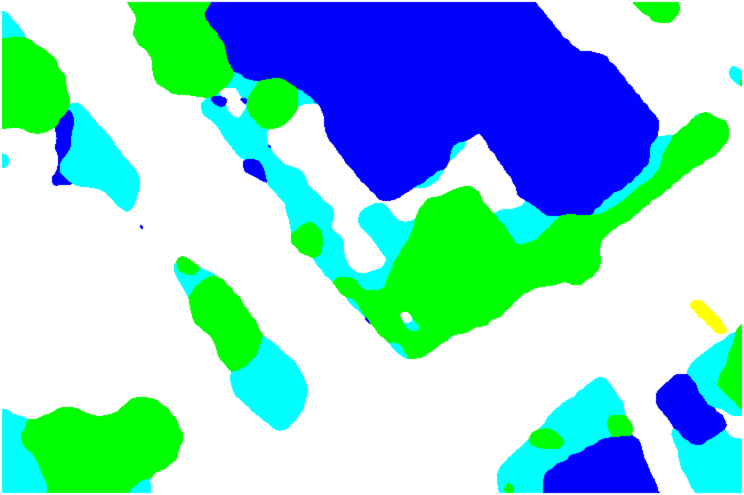} & \includegraphics[width=.13\textwidth]{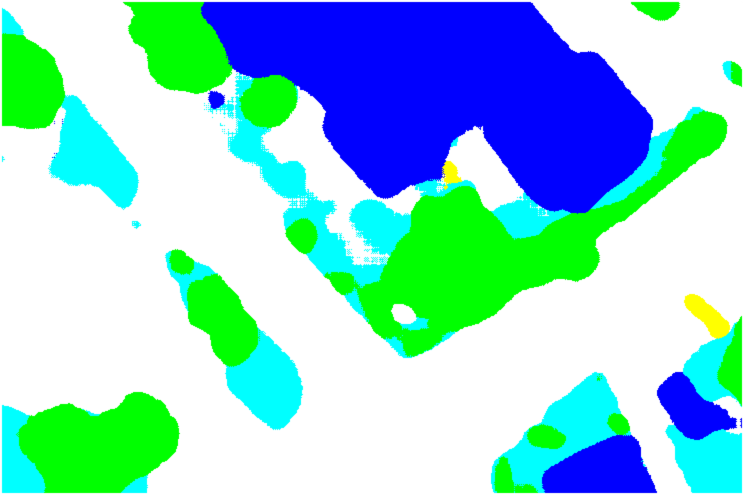} \\[-0.75mm]
%\raisebox{6mm}{\bf 2} & 
%\includegraphics[angle=90,width=.13\textwidth]{\cropdir crop-2-ti15-image.png} & 
%\includegraphics[angle=90,width=.13\textwidth]{\cropdir crop-2-ti15-ndsm.png} & \includegraphics[angle=90,width=.13\textwidth]{\cropdir crop-2-ti15-gt.png} & \includegraphics[angle=90,width=.13\textwidth]{\cropdir crop-2-ti15-msf.png} & \includegraphics[angle=90,width=.13\textwidth]{\cropdir crop-2-ti15-pc.png} & \includegraphics[angle=90,width=.13\textwidth]{\cropdir crop-2-ti15-sps.png} & \includegraphics[angle=90,width=.13\textwidth]{\cropdir crop-2-ti15-fps.png} \\
& \raisebox{15mm}{\bf 2} & 
\includegraphics[angle=0,width=.13\textwidth]{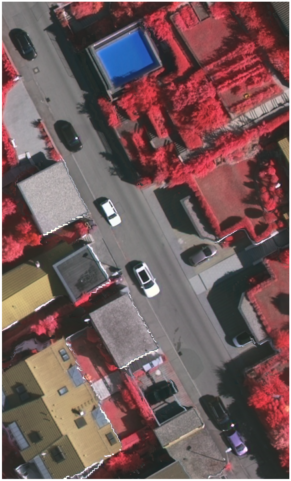} & 
\includegraphics[angle=0,width=.13\textwidth]{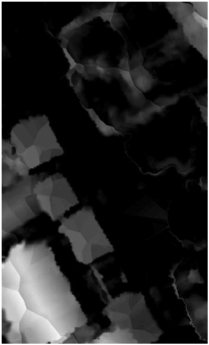} & \includegraphics[angle=0,width=.13\textwidth]{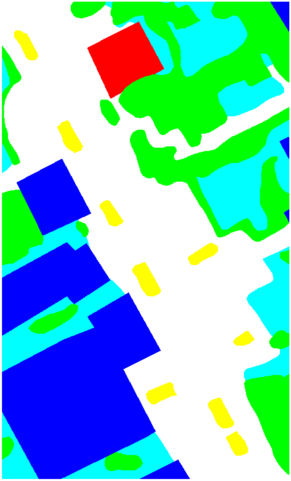} & \includegraphics[angle=0,width=.13\textwidth]{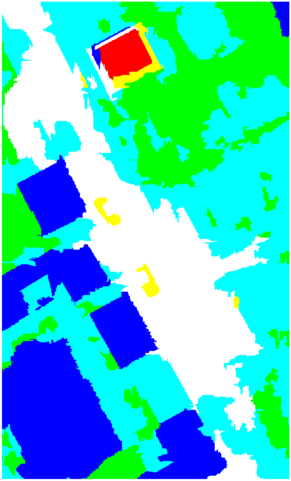} & \includegraphics[angle=0,width=.13\textwidth]{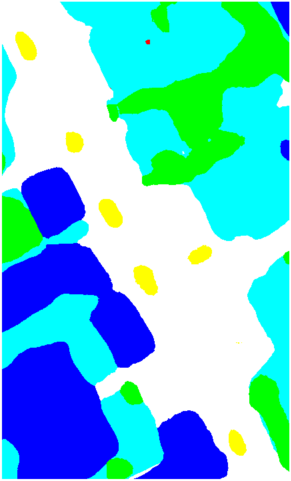} & \includegraphics[angle=0,width=.13\textwidth]{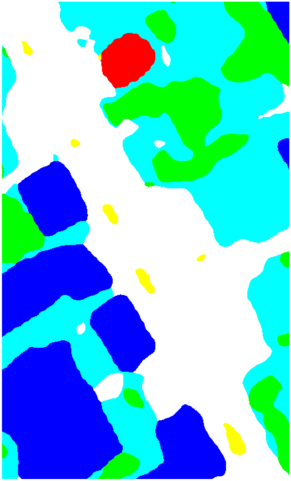} & \includegraphics[angle=0,width=.13\textwidth]{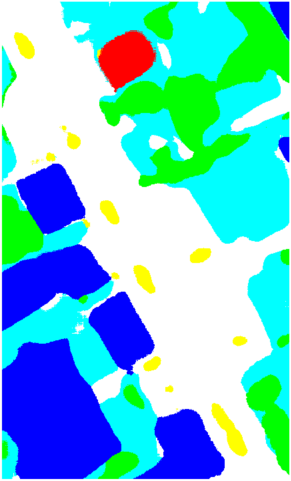} \\[-0.75mm]
& \raisebox{10mm}{\bf 3} & 
\includegraphics[width=.13\textwidth]{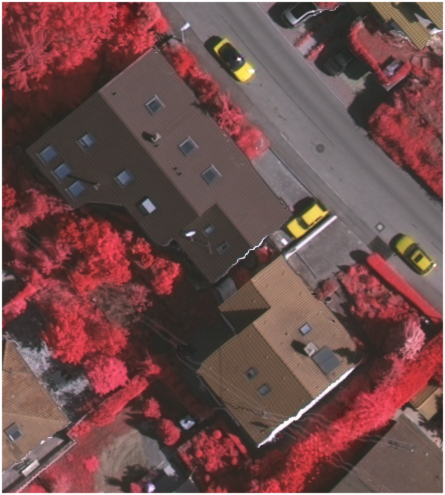} & 
\includegraphics[width=.13\textwidth]{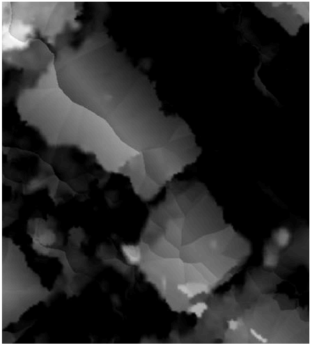} & \includegraphics[width=.13\textwidth]{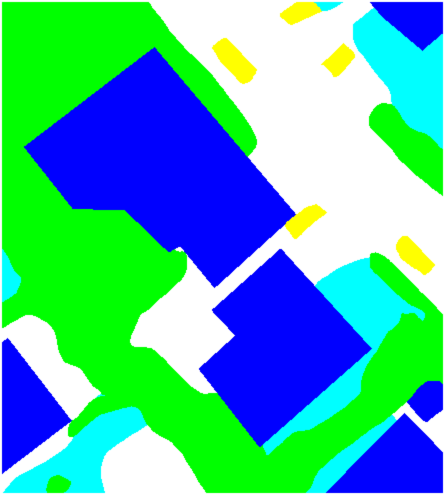} & \includegraphics[width=.13\textwidth]{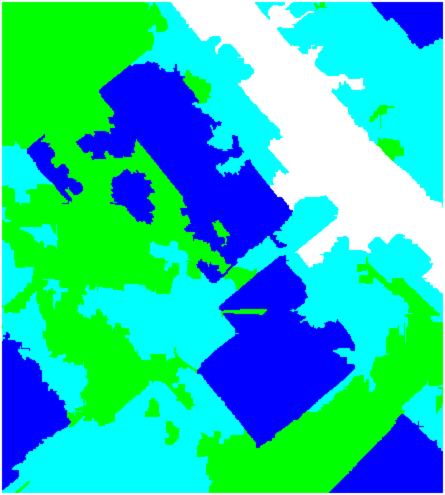} & \includegraphics[width=.13\textwidth]{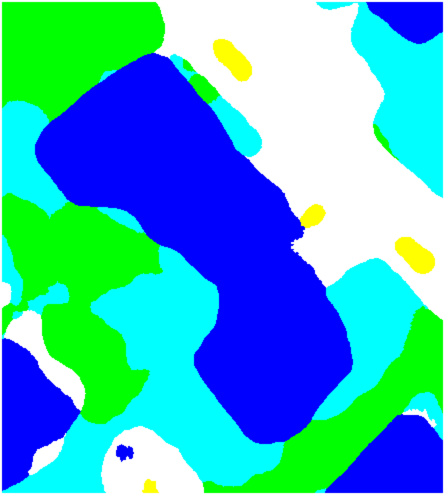} & \includegraphics[width=.13\textwidth]{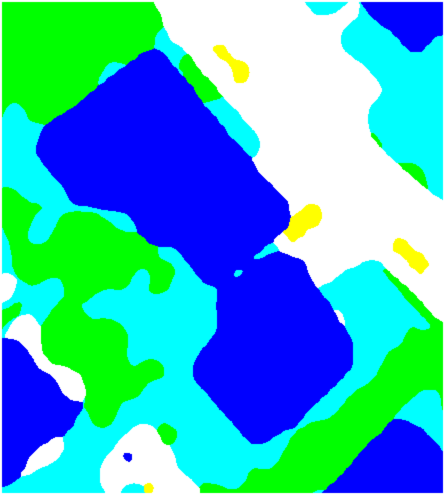} & \includegraphics[width=.13\textwidth]{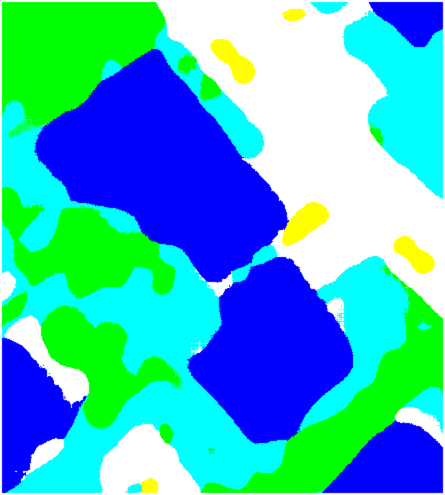} \\[-0.75mm]
%\raisebox{7mm}{\bf 4} & 
%\includegraphics[width=.13\textwidth]{\cropdirV crop-5-ti28-image.png} & 
%\includegraphics[width=.13\textwidth]{\cropdirV crop-5-ti28-ndsm.png} & \includegraphics[width=.13\textwidth]{\cropdirV crop-5-ti28-gt.png} & \includegraphics[width=.13\textwidth]{\cropdirV crop-5-ti28-msf.png} & \includegraphics[width=.13\textwidth]{\cropdirV crop-5-ti28-pc.png} & \includegraphics[width=.13\textwidth]{\cropdirV crop-5-ti28-sps.png} & \includegraphics[width=.13\textwidth]{\cropdirV crop-5-ti28-fps.png} \\
\multirow{-25}{*}{\raisebox{0cm}{\rotatebox{90}{\bf \Large  Vaihingen}}} & \raisebox{8mm}{\bf 4} & 
\includegraphics[width=.13\textwidth]{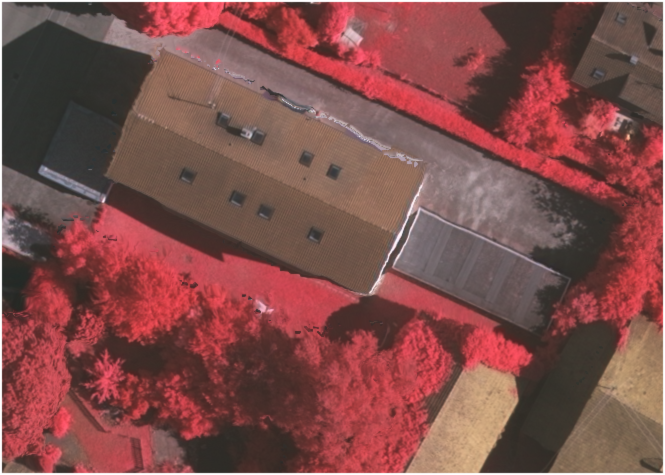} & 
\includegraphics[width=.13\textwidth]{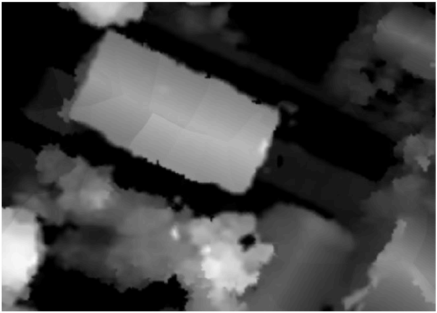} & \includegraphics[width=.13\textwidth]{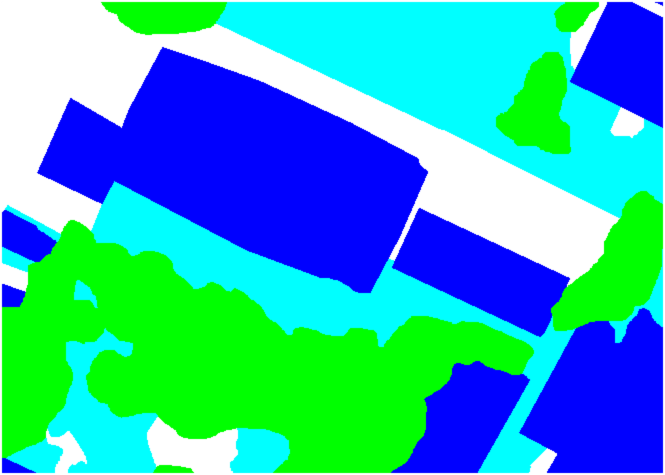} & \includegraphics[width=.13\textwidth]{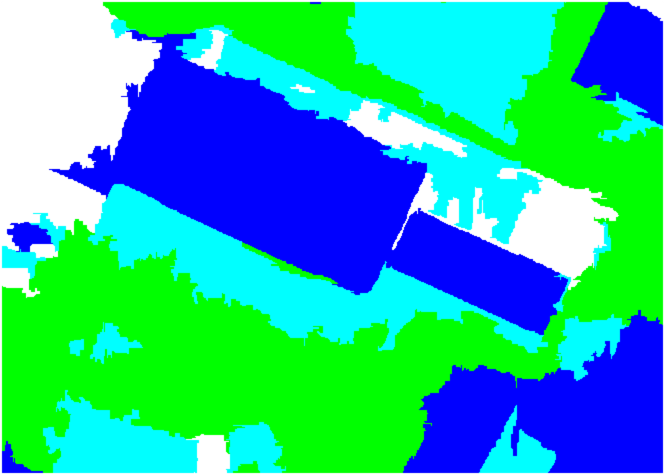} & \includegraphics[width=.13\textwidth]{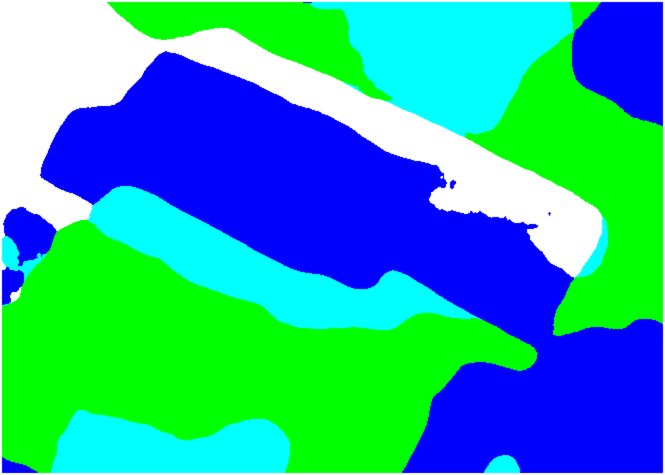} & \includegraphics[width=.13\textwidth]{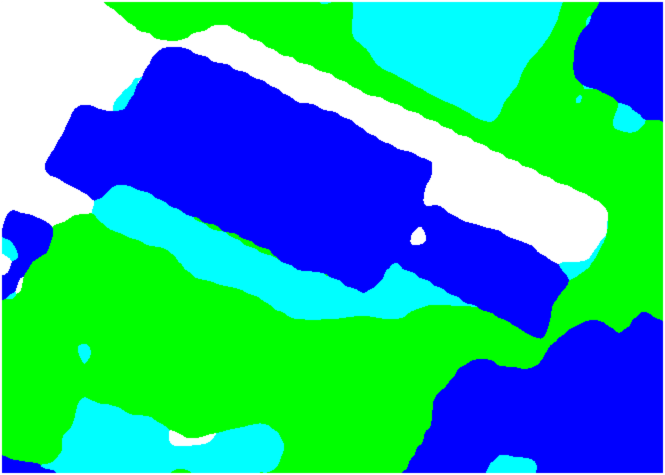} & \includegraphics[width=.13\textwidth]{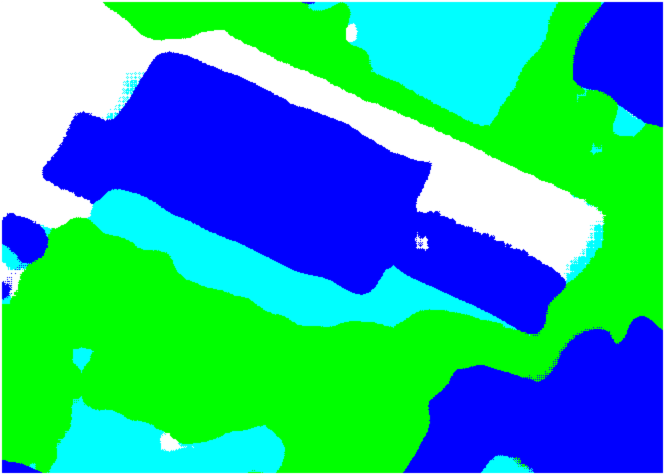} \\[-0.75mm]
\hline\hline
& \raisebox{12mm}{\bf 6} & 
\includegraphics[width=.13\textwidth]{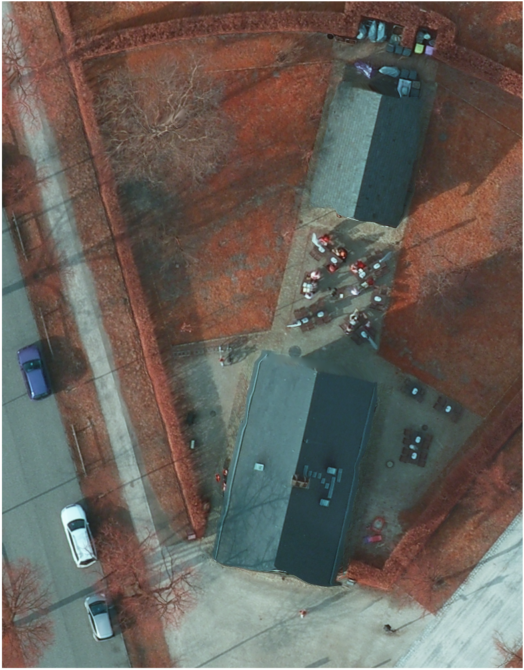} & 
\includegraphics[width=.13\textwidth]{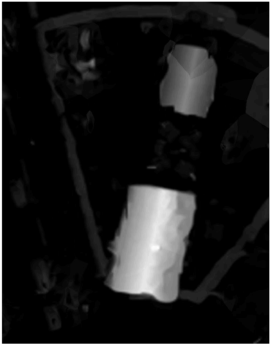} & \includegraphics[width=.13\textwidth]{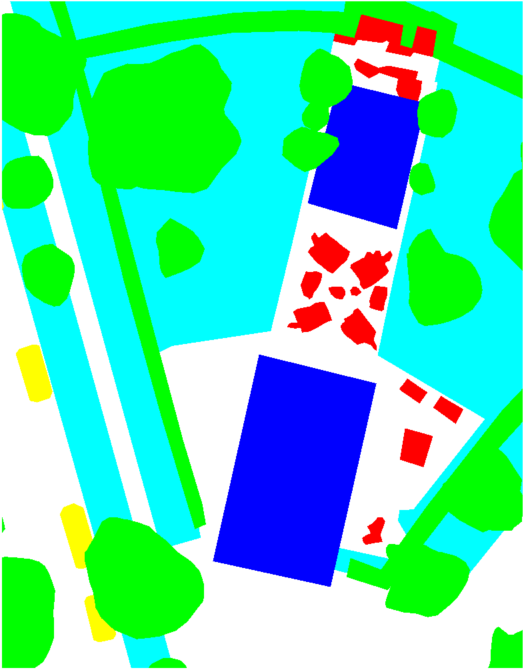} & \includegraphics[width=.13\textwidth]{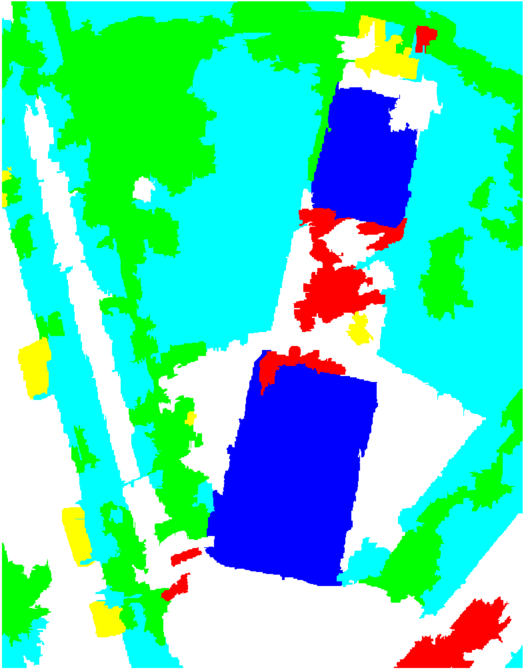} & \includegraphics[width=.13\textwidth]{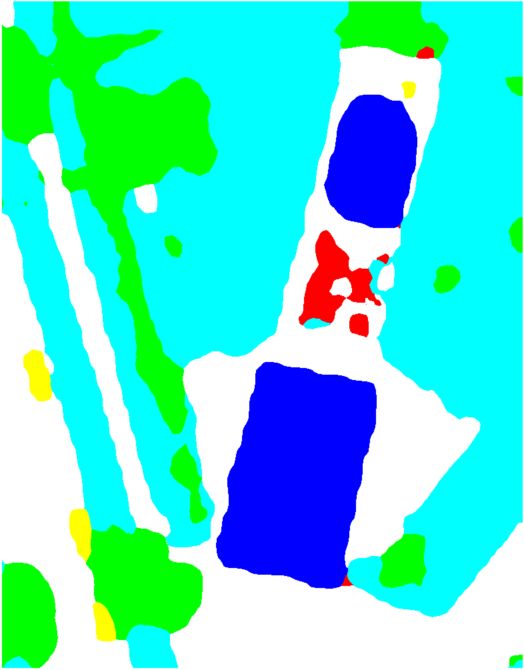} & \includegraphics[width=.13\textwidth]{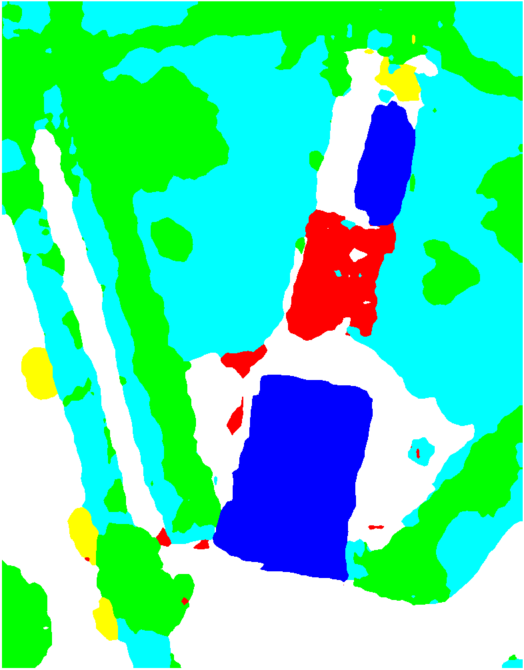} & \includegraphics[width=.13\textwidth]{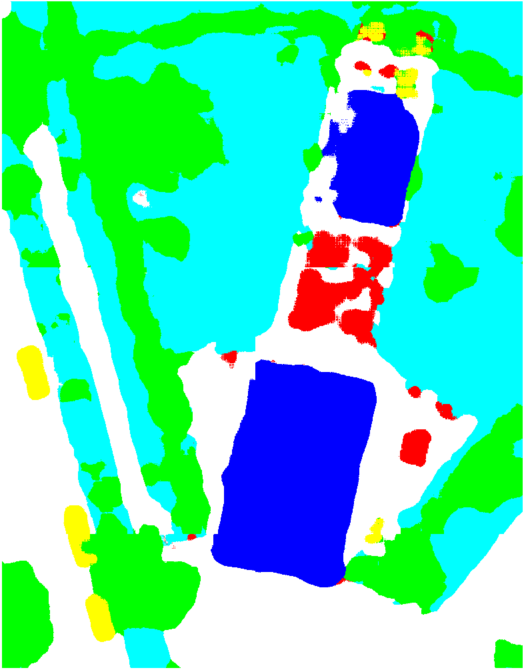} \\[-0.75mm]
& \raisebox{12mm}{\bf 7} & 
\includegraphics[width=.13\textwidth]{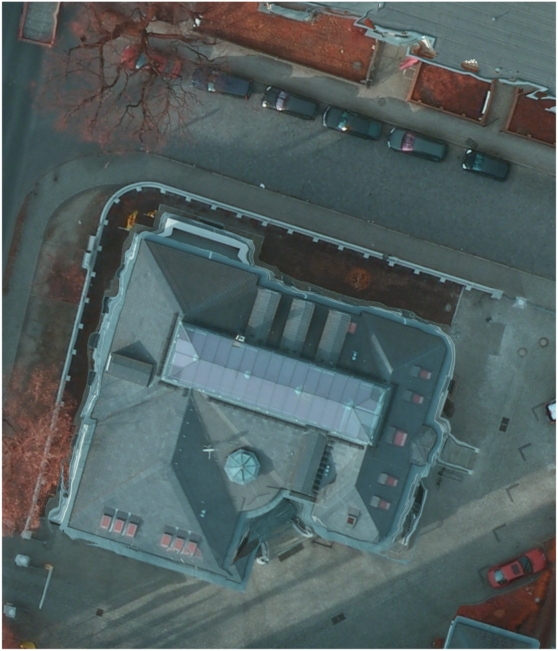} & 
\includegraphics[width=.13\textwidth]{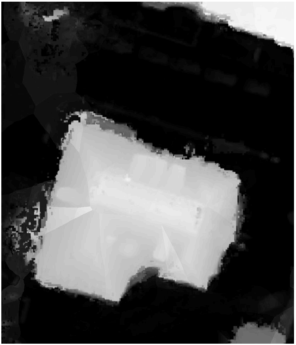} & \includegraphics[width=.13\textwidth]{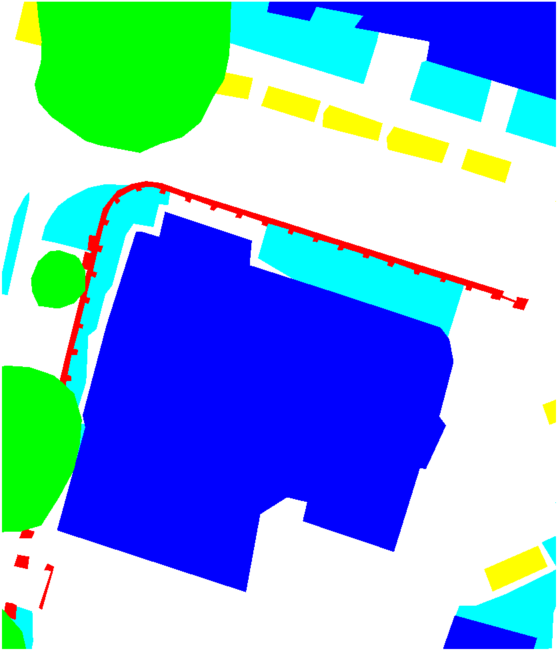} & \includegraphics[width=.13\textwidth]{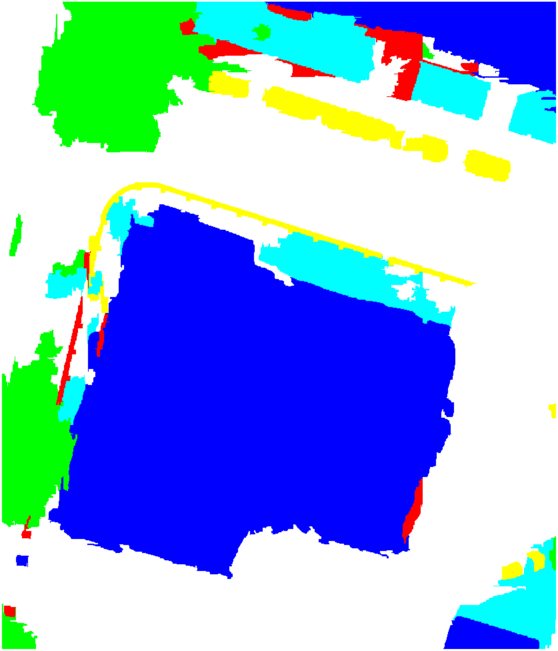} & \includegraphics[width=.13\textwidth]{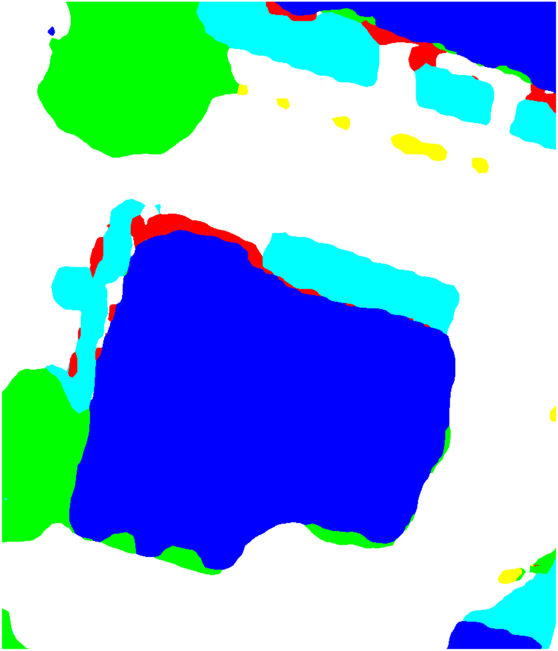} & \includegraphics[width=.13\textwidth]{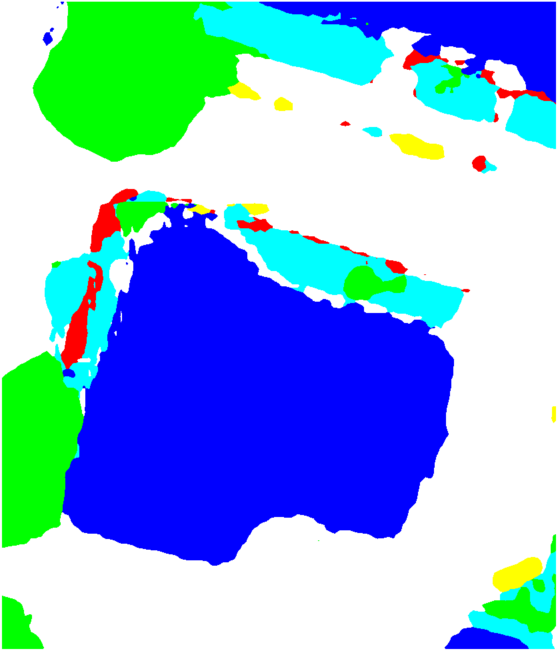} & \includegraphics[width=.13\textwidth]{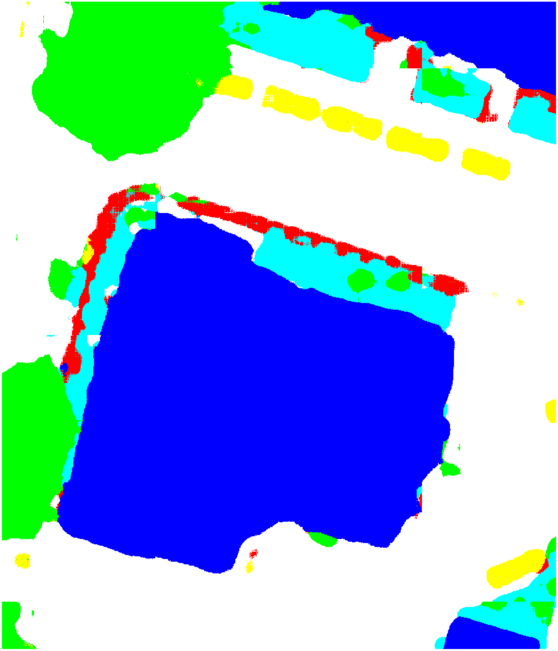} \\[-0.75mm]
& \raisebox{12mm}{\bf 8} & 
\includegraphics[width=.13\textwidth]{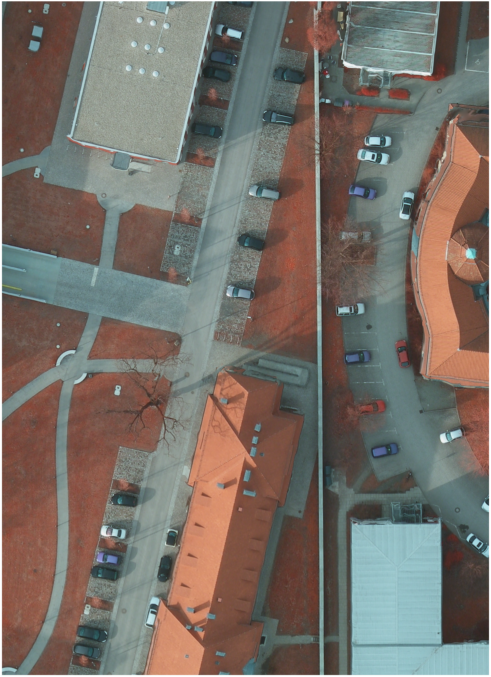} & 
\includegraphics[width=.13\textwidth]{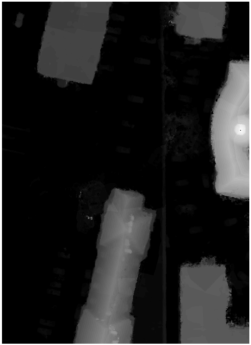} & \includegraphics[width=.13\textwidth]{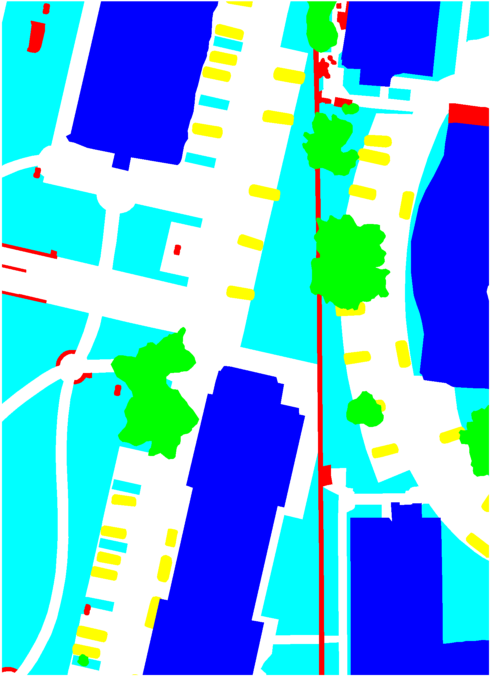} & \includegraphics[width=.13\textwidth]{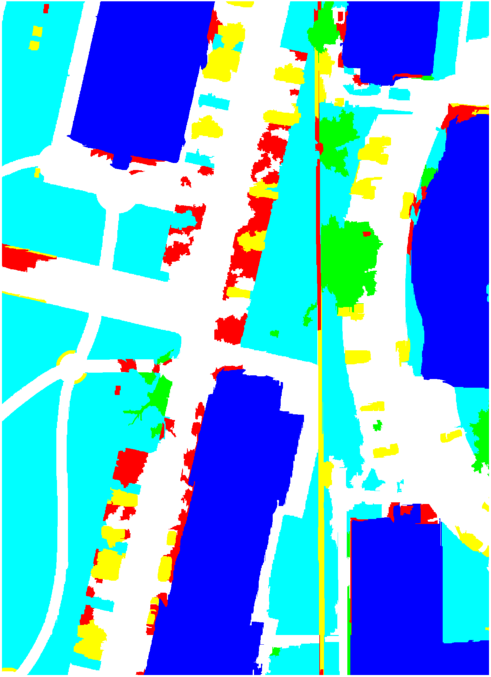} & \includegraphics[width=.13\textwidth]{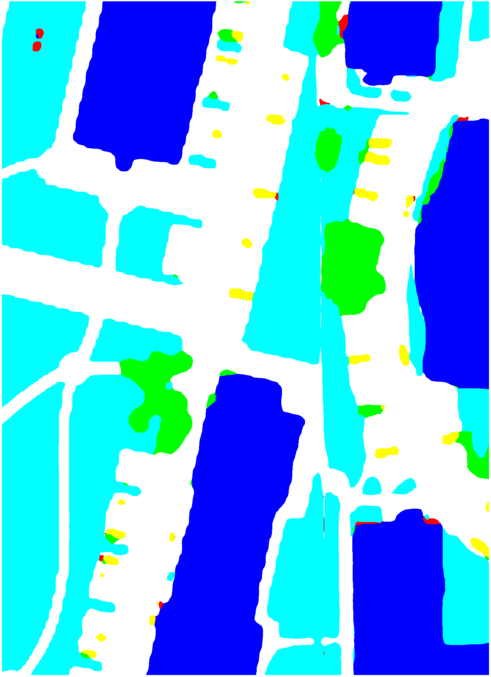} & \includegraphics[width=.13\textwidth]{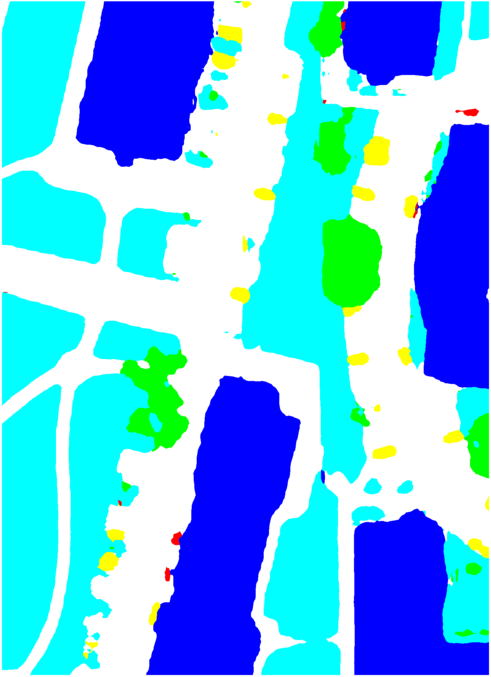} & \includegraphics[width=.13\textwidth]{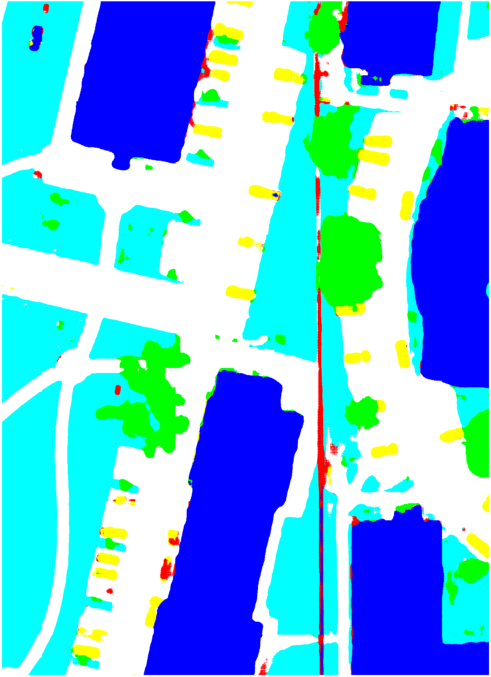} \\[-0.75mm] 
\multirow{-30}{*}{\raisebox{0cm}{\rotatebox{90}{\bf \Large  Potsdam}}} & \raisebox{12mm}{\bf 9} & 
\includegraphics[angle=90,width=.13\textwidth]{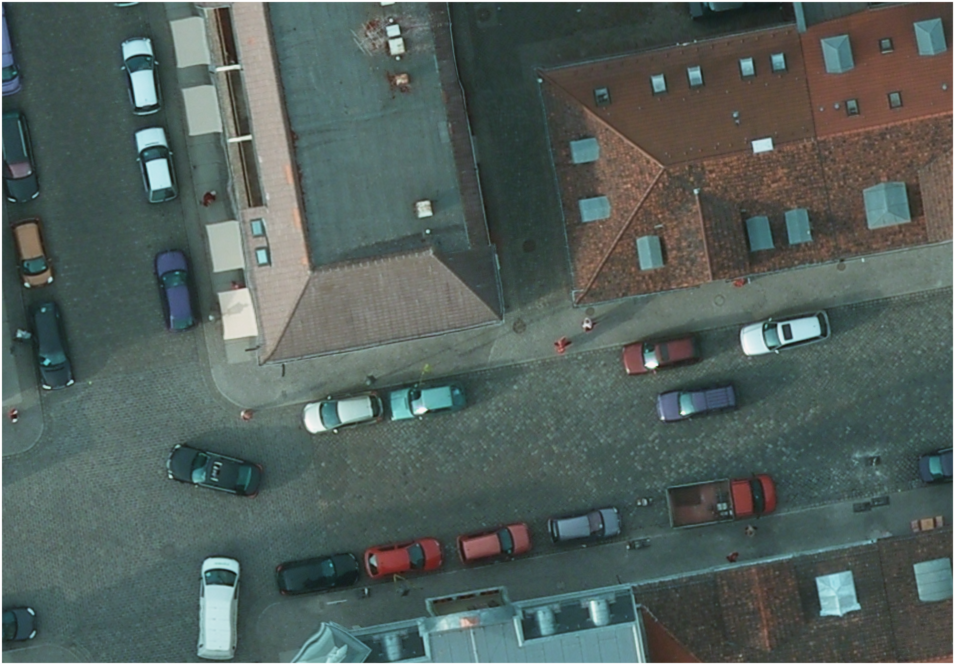} & 
\includegraphics[angle=90,width=.13\textwidth]{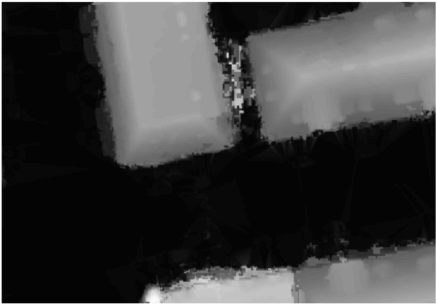} & \includegraphics[angle=90,width=.13\textwidth]{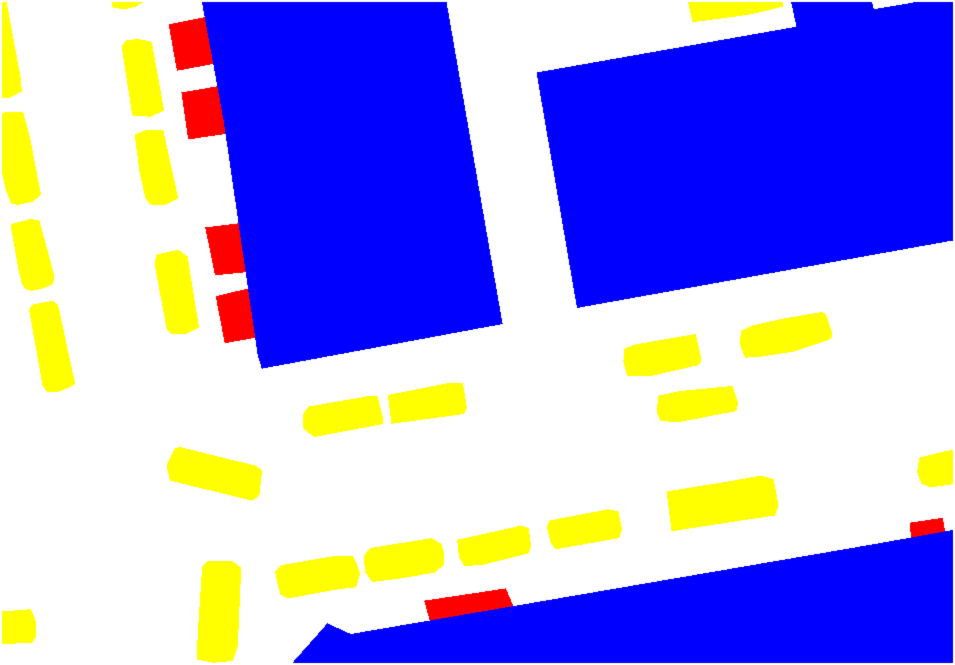} & \includegraphics[angle=90,width=.13\textwidth]{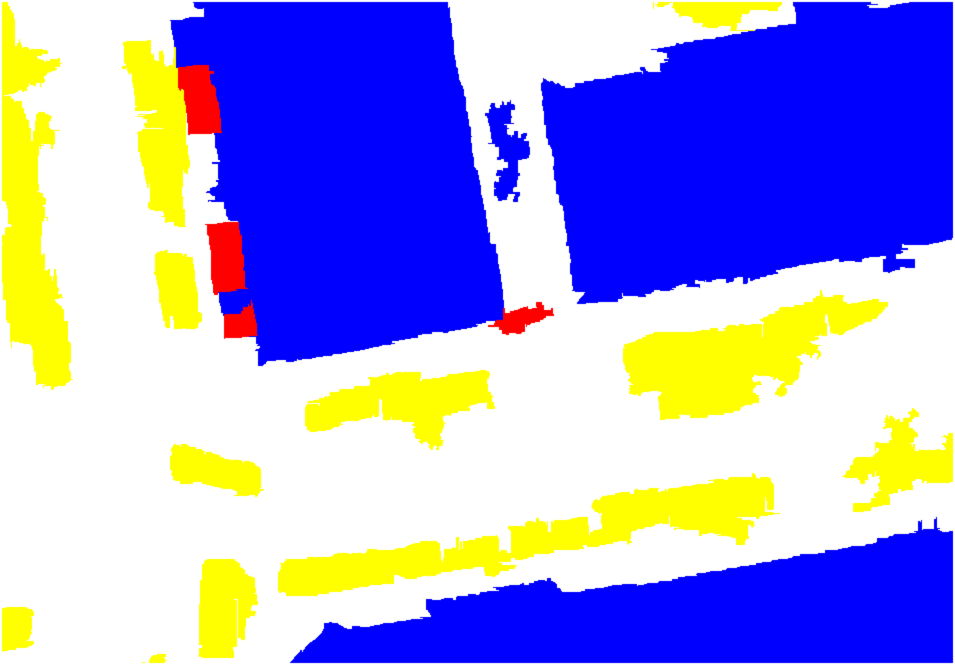} & \includegraphics[angle=90,width=.13\textwidth]{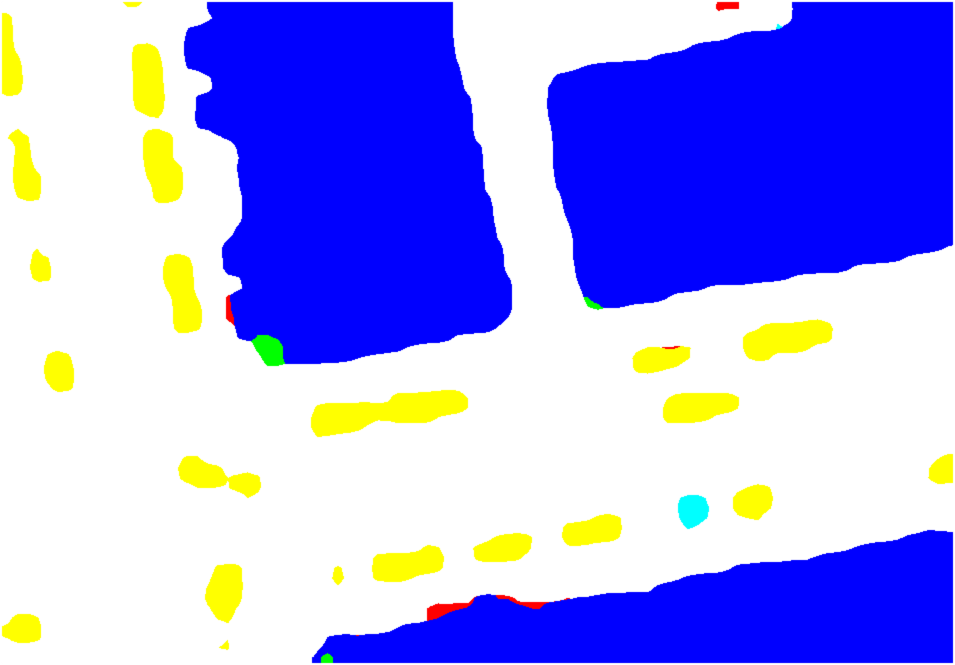} & \includegraphics[angle=90,width=.13\textwidth]{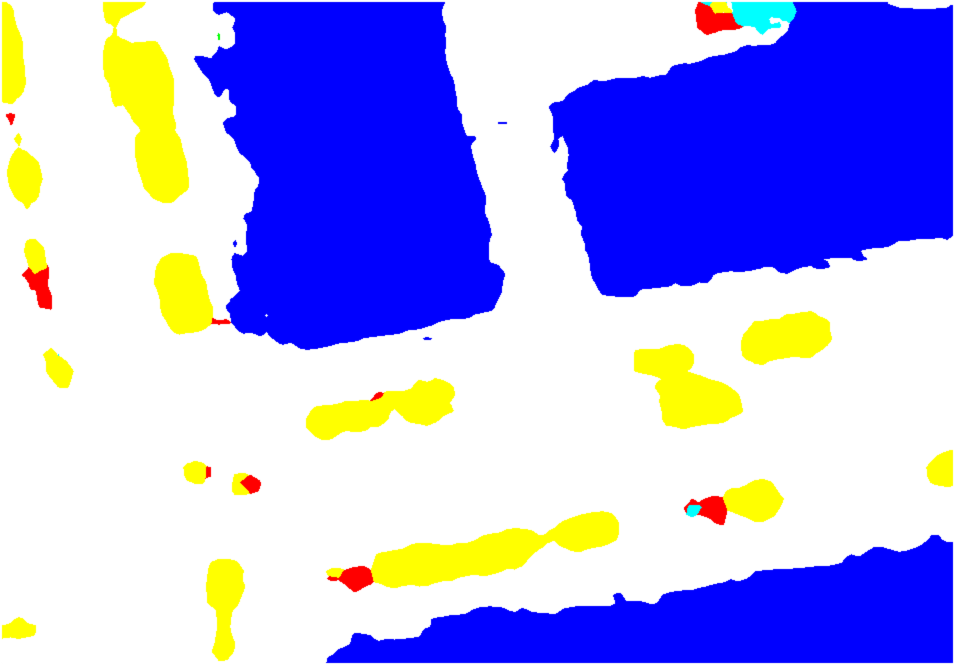} & \includegraphics[angle=90,width=.13\textwidth]{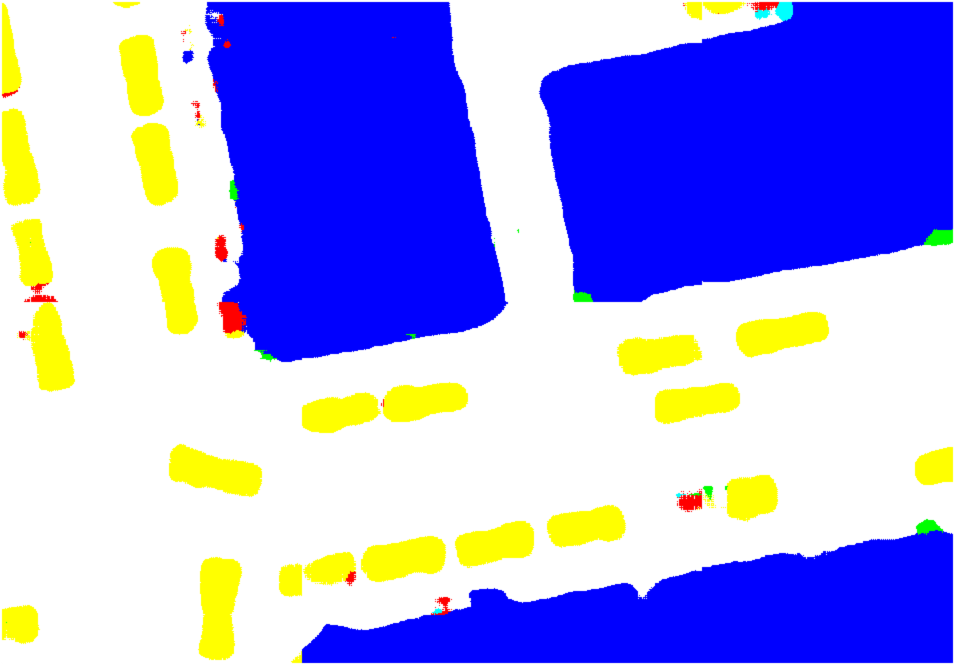} \\[-0.75mm]
\end{tabular}
\caption{Example predictions for the Vaihingen (subsets 1-5) and Potsdam (6-9) for the tested architectures. Legend -- White: impervious surfaces; \textcolor{blue}{Blue}: buildings; \textcolor{cyan}{Cyan}: low vegetation;  \textcolor{green}{Green}: trees; \textcolor{yellow}{Yellow}: cars; \textcolor{red}{Red}: clutter, background. Best viewed in color PDF. \label{fig:examplepredictions}}
\end{figure*}

\begin{figure*}
\centering
\begin{tabular}{ccc|cccc}
{\bf Image} & {\bf nDSM} & {\bf GT} & {\bf SP-MSF} & {\bf CNN-PC} & {\bf CNN-SPL} & {\bf CNN-FPL} \\ \hline
\includegraphics[width=.13\textwidth]{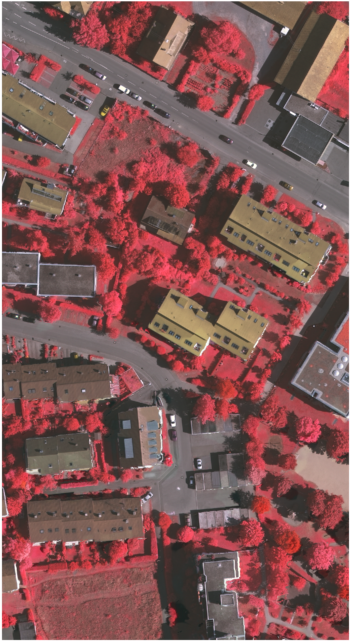} & 
\includegraphics[width=.13\textwidth]{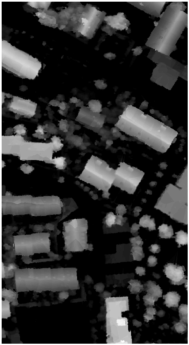} & \includegraphics[width=.13\textwidth]{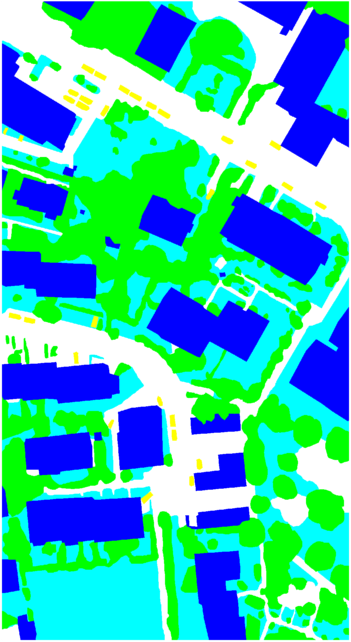} & \includegraphics[width=.13\textwidth]{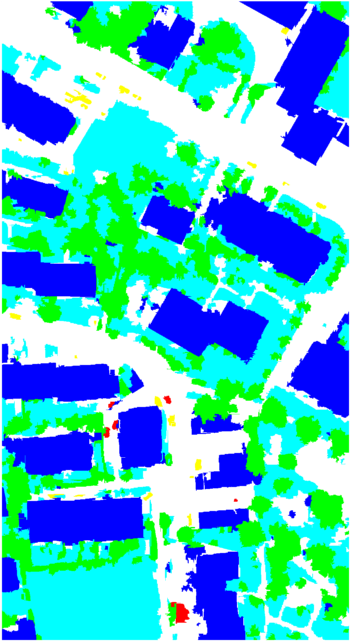} & \includegraphics[width=.13\textwidth]{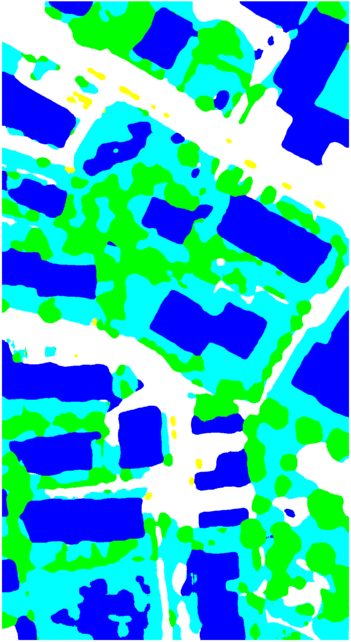} & \includegraphics[width=.13\textwidth]{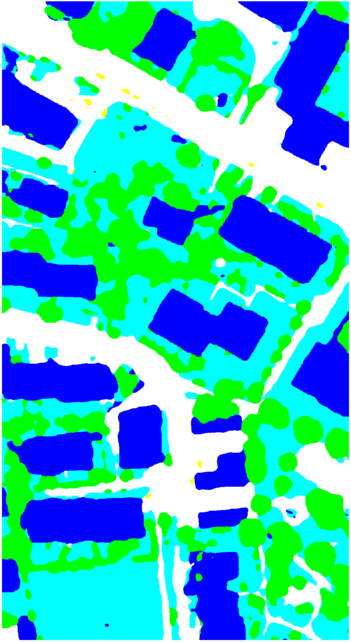} & \includegraphics[width=.13\textwidth]{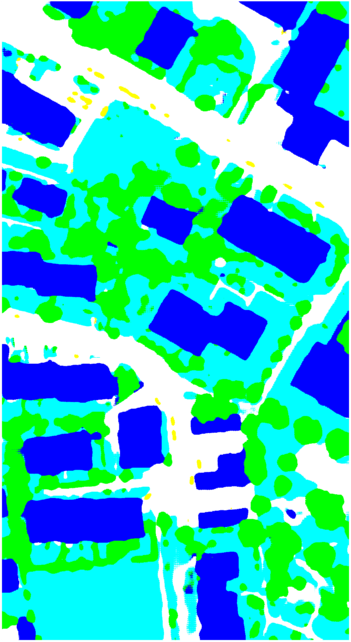} \\
\end{tabular}
\caption{Full prediction for tile ID 34. Legend -- White: impervious surface; \textcolor{blue}{Blue}: buildings; \textcolor{cyan}{Cyan}: low vegetation;  \textcolor{green}{Green}: trees; \textcolor{yellow}{Yellow}: cars; \textcolor{red}{Red}: clutter.\label{fig:fullpred34}}
\vspace{-4mm}
\end{figure*}

In Fig.~\ref{fig:examplepredictions}(1-4) we summarize some aspects of the predictions by clipping map portions from 4 out of the 5 validation images (the 5th image is shown in its entirety in Fig.~\ref{fig:fullpred34}). %From left to right, the array of figure shows the identifier of the clip, the image, the {\it n}DSM, the ground truth (GT), the superpixel baseline (SP-MSF) and the three CNN architecture tested -- CNN-PC, CNN-SPL and CNN-FPL. 
In all the clips, we can see how the different methods act on boundaries, in particular for the class ``building''. For buildings having high contrasted boundaries, SP-MSF provides the best preservation of their geometry. However, there are many cases of ``building'' instances and particularly for other classes, in which boundaries are not sharp: in these cases SP-MSF is prone to fail. For single regions with ambiguous appearance, the predictions can be noisy and, since the context of each region is not taken into account, can result in wrong assignments. CNN-based strategies result in more accurate and semantically coherent segmentations. CNN-PC, because of its unstructured nature, does not preserve well object boundaries and tends to overmooth classes with complex boundaries. As expected, CNN-SPL offers a trade-off between CNN-PC and CNN-FPL, which, in turn, offers best segmentations. CNN-FPL makes better use of class co-occurrences, for instance by avoiding spurious prediction of small patches of the class building (e.g. in clip 1). CNN-FPL deals better with thin and elongated elements and boundaries in general, by preserving the shape of such structures (e.g. building shape in clip 4 and the gap between the buildings in clip 3) thanks to the learned upsamplings.

The class ``car'' is very difficult to correctly segment. SP-MSF often misclassifies ``cars'' because superpixels do not always isolate class instances. Again, CNNs are generally more accurate. CNN-PC, although showing good segmentation for detected cars, misses most of them. Semantic labeling CNN are more accurate in detecting single cars, and, in particular CNN-FPL offers a good trade-off between segmentation accuracy and detection (e.g. clips 2 and 3). The class ``background'' is detected by all methods with different success rates, mostly depending on its local appearance (recall that this class is not semantically nor visually coherent, since collecting different semantic classes).

Coupling elevation and spectral information eases the detection of buildings and trees (the elevated classes). Such information obviously helps in discriminating ambiguous occurrences of the class building, e.g. rooftops showing the same appearance as roads. %For instance, all the models agree on detecting a building in the upper part of clip 8, which, by looking at the {\it n}DSM, seems to be an error in the GT. 
Rooftops covered in dense vegetation are often misclassified as trees and never as grass, since being on two different elevation levels. In Fig.~\ref{fig:fullpred34} the full segmentation of image tile 34 is given. What observed for the clips above summarizes the classification of the entire tile. 

\renewcommand{\tabcolsep}{1pt}

\renewcommand{\tabcolsep}{1pt}

% Filters learned at the first layer are shown in Fig.~\ref{fig:learnedfilters}. The three different architectures learned basic image aspects, such as color gradient, line detectors and Gaussian-like smoothings. For the CNN-PC, such filters look visually similar to published works dealing with image classification, underlying once more the generality of filters learned by the first layer. For the CNN-SPS and CNN-FPS, some filters look slightly different being trained for a different task. For instance, some filters have grid-like structures, probably reacting differently to the texture of different classes. 

% \begin{figure*}
% \centering
% \renewcommand{\tabcolsep}{10pt}
% \begin{tabular}{ccc}
% \includegraphics[width=.2\textwidth]{\figdir cnn1x1filters.pdf} & 
% \includegraphics[width=.2\textwidth]{\figdir cnnPofilters.pdf} & 
% \includegraphics[width=.2\textwidth]{\figdir cnnFdfilters.pdf} \\
% (a) & (b) & (c)\\ 
% \end{tabular}
% \caption{Filters learned at the first convolutional layer for the different CNN architectures: (a) CNN-PC, (b) CNN-SPS and (c) CNN-FPS. \label{fig:learnedfilters}}
% \end{figure*}

\subsubsection{Submission to challenge}

To compare to state-of-the-art models, we submitted the maps obtained for the 17 unlabeled tiles, to obtain the test accuracy. With the independent Vaihingen evaluation criteria, we scored 87.3 points in overall accuracy. This sets the CNN-FPL as the 5th most accurate model\footnote{http://www2.isprs.org/vaihingen-2d-semantic-labeling-contest.html, submission results as March the 29th, 2016 (UZ\_1 entry).}, on a tie with the ADL\_2 entry \cite{paisitkriangkrai2015cvprw}. CNN employed by other participants ranged from fully convolutional networks \cite{long2015cvpr} to ensemble of (multiscale) patch-based CNN. However, it is worth pointing out that all the models showing higher accuracy on this test set combine either features learned by CNN and handcrafted features, classified by an additional external nonlinear classifier (e.g. random forests) and / or post-processed by a conditional random field of varying complexity. In our setting, we \emph{only} employ classification scores as given by CNN-FPL, with \emph{no additional postprocessing techniques}. We could argue that by adding manually extracted features (to compensate for appearance variations not learned by the CNN) and / or adding a further smoothing by random fields models, we could gain a few additional accuracy points, as pointed out in \cite{sherrah2016arxiv}. But this would hinder the contribution of this paper.

\subsection{Potsdam dataset results}

\begin{table}
\renewcommand{\tabcolsep}{3pt}
\caption{Numerical results for the Potsdam validation set. %We report figures of merit accounting for the background class ({\bf full}) and without it ({\bf no bk}), as well as on by employing the ground truths with eroded edges, on both background ({\bf er full}) and no-background cases ({\bf er no bk}).
\label{tab:potdsamresults}}
\centering
\begin{tabular}{c|c|cccc}
\toprule
  & {\bf Model}   & {\bf OA}    & {\bf K}     & {\bf AA}    & {\bf F1} \\ \hline 
  & SP-MSF        & 82.29       & 76.28       & 72.27       & 68.55    \\ 
  & CNN-PC        & 84.07       & 78.13       & 63.67       & 66.83    \\ 
  & CNN-SPS       & 83.00       & 76.93       & 66.91       & 67.90    \\ 
 \raisebox{1mm}{\multirow{-3}{*}{\rotatebox{90}{\bf full}}}   
  & CNN-FPS       & {\bf 85.85} & {\bf 80.88} & {\bf 74.31} & {\bf 73.81} \\
\midrule
 % & model  & OA    & K     & AA    & F1    & t [s]  \\\hline
  & SP-MSF        & 85.73       & 80.55       & 83.24       & 80.16    \\ 
  & CNN-PC        & 85.83       & 80.36       & 74.27       & 77.33    \\
  & CNN-SPS       & 84.94       & 79.35       & 78.24       & 78.85    \\ 
 \raisebox{1mm}{\multirow{-3}{*}{\rotatebox{90}{\bf no bk}}}   
  & CNN-FPS       & {\bf 87.94} & {\bf 83.52} & {\bf  86.06} & {\bf 85.32} \\
\midrule
 % & m & OA    & K     & AA    & F1    & t [s]  \\
  & SP-MSF        & 84.40       & 78.94       & 74.89       & 70.30    \\ 
  & CNN-PC        & 86.58       & 81.47       & 66.35       & 69.78    \\ 
  & CNN-SPS       & 85.49       & 80.18       & 69.49       & 70.51    \\
 \raisebox{1mm}{\multirow{-3}{*}{\rotatebox{90}{\bf er full}}}
  & CNN-FPS       & {\bf 88.04} & {\bf 83.72} & {\bf  76.82} & {\bf 76.12} \\
\midrule
 % & m & OA    & K     & AA    & F1    & t [s]  \\
  & SP-MSF        & 87.56       & 82.94       & 85.75       & 82.16    \\ 
  & CNN-PC        & 88.05       & 83.36       & 77.26       & 80.52    \\
  & CNN-SPS       & 87.14       & 82.30       & 81.01       & 81.64    \\
 \raisebox{2mm}{\multirow{-3}{*}{\rotatebox{90}{\bf er no bk}}}   
  & CNN-FPS       & {\bf 89.86} & {\bf 86.06} & {\bf 88.79} & {\bf 87.97} \\
\bottomrule
\end{tabular}
\vspace{-3mm}
\end{table}

For this dataset, we trained the CNN models in two steps: In the first step, we created the super-batch by sampling all the 65$\times$65 patches with an overlap of 33 pixels, for each image at a time. We iterated over all the entire training images for 200 epochs (i.e. no class-balanced random sampling). After that, we used the strategy employed for the Vaihingen dataset, which samples patches approximately uniformly across classes. We employed this strategy to speed up training by first learning the most important recurring patterns / classes in the images. The second step aimed at fine tuning the network to learn specific, finer class-appearance representations.

\subsubsection{Numerical results}

In Tab.~\ref{tab:potdsamresults} we present results using the validation images presented in Section~\ref{sec:data}. For the CNN-PC approach we performed inference with a stride of 5, since the size of the images is significantly larger (6000 $\times$ 6000 pixels) and the spatial resolution of the images is roughly the half (5cm). The loss in accuracy when using a stride of 5 is less than 0.5\%, but the computational time is immensely reduced (the system 1'440'000 windows instead of 36'000'000, so 25$\times$ less). We use bilinear upsampling on the posterior probability maps to upsample results to the original image size, similar to the last layer of \cite{long2015cvpr}.

As observed for the Vaihingen dataset, the superpixel baseline offers more balanced errors across classes when compared to the CNN-PC, since the CNN-PC shows higher OA and K scores while lower AA and F1. Both approaches perform similarly on the ``clutter'' class, since balanced evaluation metrics significantly increase while removing such class. Directly predicting patches with the CNN-SPL strategy results in global metrics roughly on par to the baselines SP-MSF and CNN-PC, while slightly better than CNN-PC on balanced metric. As for the Vaihingen dataset, the improved modeling power of the CNN-FPL offers better accuracies for all the accuracy metrics considered.

\subsubsection{Qualitative results}

In Fig.~\ref{fig:examplepredictions}(6-9) we plot examples of semantic labelings on subsets of the validation images. Due to the higher spatial resolution (5cm instead of 9cm for the Vaihingen dataset) the land-cover classes are represented by a more variate appearance, in both color and size. For instance, in clips 7 and 8 the thin fences and wall correspond to class ``clutter'' and only SP-MSF and CNN-FPS are able to detect it, while only the latter scoring the correct class most of the time. We acknowledge the good performance of the SP-MSF, but we also note that ambiguous appearance of superpixels cannot be solved. Cars and buildings are generally segmented correctly by all the CNNs, but only the CNN-FPS is able to segment them in a geometrically accurate manner, e.g. single cars segmented as whole objects and not as multiple parts or undersegmented ones (CNN-PC and CNN-SPL). Again, this beneficial effect stems form learned deconvolutions, to upsample to full image resolution. We also note that for this dataset, the class ``clutter'' and ``trees'' are hard to model. The former situation is mostly due to the very variate and mixed nature of it. For the latter, the fact that some trees do not have leaves makes the modeling of their actual extent hard, or even hard to detect when they stand on grass (see e.g. clips 6 and 7). CNN-based system are able to perform well, but again SP-MSF misses the difficult instances.

\subsubsection{Submission to challenge}

To compare to recent submissions of \cite{sherrah2016arxiv}, we submitted the prediction for the 14 test images to the Potsdam 2D semantic labeling challenge\footnote{http://www2.isprs.org/potsdam-2d-semantic-labeling.html, submission results as August 2nd, 2016 (UZ\_1 entry).}. Test maps show and 85.8\% overall accuracy, roughly 5\% less accurate than the approaches in \cite{sherrah2016arxiv}. In particular, the class ``tree'' is around 7-8\% worse than the aforementioned entry. However, we believe results are satisfactory, in particular since we are using significantly simpler network architectures if compared to the models presented in \cite{simonyan2015iclr} (VGG16) and employing no post-processing, which could potentially lead to better results.

%%% Local Variables: 
%%% mode: latex
%%% TeX-master: "main-volpi-cnn-deconv-journal"
%%% End:

%\section{Discussion}\label{sec:disc}

\section{Discussion and Conclusion}\label{sec:concs}

In this paper, we presented an approach to perform dense semantic labeling using convolutional neural networks. The adopted architecture first encodes concepts in a rough spatial map represented by many channels, and then learns rules to upsample this spatially coarse features back to the original resolution, via learned deconvolutions. We proposed this system to cope with the high spatial and geometrical information contained in ultra-high resolution images ($<$ 10cm), usually coupled to little spectral information. However, the full patch segmentation network can be also applied to scenarios, where normalized digital surface models are not available, spectral channels are numerous and resolution is coarser. The main challenge to transfer such approach to the processing of satellite images would be the availability of densely annotated ground truth, to train discriminatively CNN models.

Numerical and qualitative results illustrated the advantages of learning CNN directly for segmentation tasks, as underlined in \cite{sherrah2016arxiv} as well. Predicting the segmentation for full patches is actually advantageous from both the efficiency and semantic / geometric accuracy perspectives. Practically, if compared to standard CNN performing patch classification (i.e. CNN-PC) we were able not only to leverage the power of semantic abstraction of standard CNN, but we could also learn nonlinear upsamplings, encoding class-relationships and co-occurrences at a higher semantic level. This is a direct consequence of interdepentent predictions, thanks to the hierarchical downsample-then-upsample architecture. One can interpret the learned upsamplings as activation specific interpolation filters, encoding the spatial dependence of locations. Advantages are clear when comparing to models predicting each location in isolation, based on the appearance of the patch. 

We obtained results aligned with the state-of-the-art models on two extremely challenging datasets, without performing any post-processing (e.g. CRF or MRF) and without recurring to strategies involving external classifiers and additional hand-crafted features.

%%% Local Variables:
%%% mode: latex
%%% TeX-master: "main-volpi-cnn-deconv-journal"
%%% End:

\section*{Acknowledgments}
This work was supported in part by the Swiss National Science Foundation, via the
  grant 150593 ``Multimodal machine learning for remote sensing information fusion'' (http://p3.snf.ch/project-150593).

\bibliographystyle{IEEEtran}

\vfill 

\end{document}